\documentclass[times,twocolumn,final]{elsarticle}

\usepackage{medima}
\usepackage{framed,multirow}

\usepackage{amssymb}
\usepackage{latexsym}

\usepackage{url}
\usepackage{xcolor}

\usepackage{hyperref}

\definecolor{newcolor}{rgb}{.8,.349,.1}

\journal{Medical Image Analysis}
\usepackage{amsmath}
\usepackage{amsthm}

\usepackage{bm}
\usepackage{breqn} 

\usepackage{hhline} 

\usepackage{subcaption} 

\usepackage{pdfpages} 
\newtheorem{theorem}{Theorem}

\begin{document}

\verso{David Wiesner \textit{et~al.}}

\begin{frontmatter}

\title{Generative modeling of living cells with SO(3)-equivariant implicit neural representations}
 
\author[1]{David \snm{Wiesner}\corref{cor1}}
\cortext[cor1]{Corresponding author: \textit{e-mail:} \texttt{wiesner@fi.muni.cz} }
\author[2]{Julian \snm{Suk}}
\author[2]{Sven \snm{Dummer}}
\author[1]{Tereza \snm{Ne\v{c}asov\'{a}}}
\author[3]{Vladim\'ir \snm{Ulman}}
\author[1]{David \snm{Svoboda}}
\author[2]{Jelmer M. \snm{Wolterink}}

\address[1]{Centre for Biomedical Image Analysis, Masaryk University, Brno, Czech Republic}
\address[2]{Department of Applied Mathematics \& Technical Medical Centre, University of Twente, Enschede, The Netherlands}
\address[3]{IT4Innovations, VSB -- Technical University of Ostrava, Ostrava, Czech Republic}

\received{---}
\finalform{---}
\accepted{---}
\availableonline{---}
\communicated{---}

\begin{abstract}
Data-driven cell tracking and segmentation methods in biomedical imaging require diverse and information-rich training data. In cases where the number of training samples is limited, synthetic computer-generated data sets can be used to improve these methods. This requires the synthesis of cell shapes as well as corresponding microscopy images using generative models. To synthesize realistic living cell shapes, the shape representation used by the generative model should be able to accurately represent fine details and changes in topology, which are common in cells. These requirements are not met by 3D voxel masks, which are restricted in resolution, and polygon meshes, which do not easily model processes like cell growth and mitosis. 
In this work, we propose to represent living cell shapes as level sets of signed distance functions (SDFs) which are estimated by neural networks. We optimize a fully-connected neural network to provide an implicit representation of the SDF value at any point in a 3D+time domain, conditioned on a learned latent code that is disentangled from the rotation of the cell shape. We demonstrate the effectiveness of this approach on cells that exhibit rapid deformations (\textit{Platynereis dumerilii}), cells that grow and divide (\textit{C. elegans}), and cells that have growing and branching filopodial protrusions (A549 human lung carcinoma cells). 
A quantitative evaluation using shape features and Dice similarity coefficients of real and synthetic cell shapes shows that our model can generate topologically plausible complex cell shapes in 3D+time with high similarity to real living cell shapes. Finally, we show how microscopy images of living cells that correspond to our generated cell shapes can be synthesized using an image-to-image model.

\vspace{6mm}

\end{abstract}


\end{frontmatter}

\section{Introduction}
\label{intro}
Accurate and reliable segmentation of biomedical optical microscopy images is a challenging task~\citep{meijering2012cell,meijering2020bird}, which is extremely time-consuming and tedious when performed manually, especially on 3D and 3D+time data ~\citep{coutu2013probing,webb2003assessing}. There is a clear need for automatic segmentation methods, and deep learning methods have shown promising results~\citep{stringer2021cellpose}. However, these methods require large sets of information-rich and varied training data~\citep{dataavailability}, consisting of pairs of microscopy images and their target segmentation masks.

\begin{figure*}[!t]
\centering

\hfill
\includegraphics[width=0.986\textwidth]{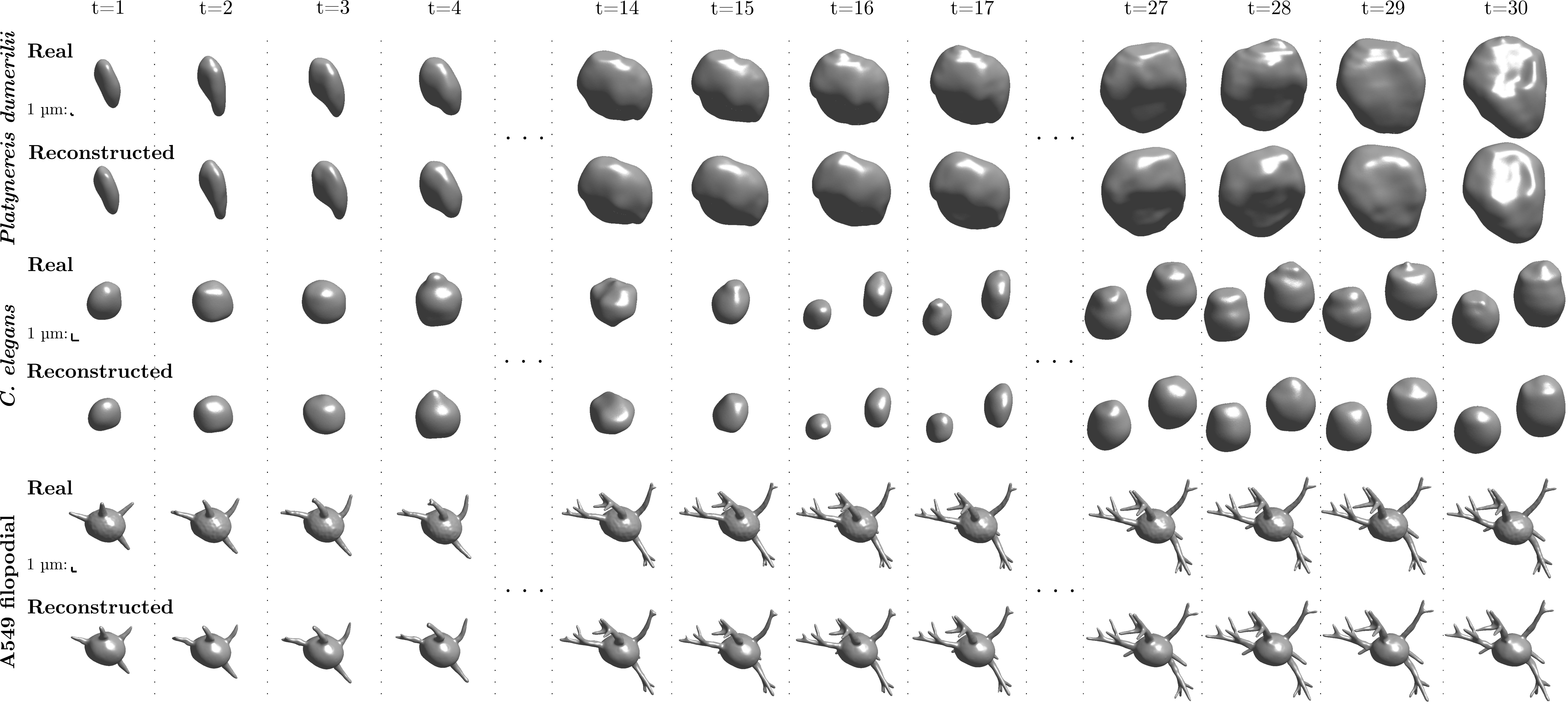}

\footnotesize{(a) Real and reconstructed cell shapes}

\vspace{2mm}

\includegraphics[width=\textwidth]{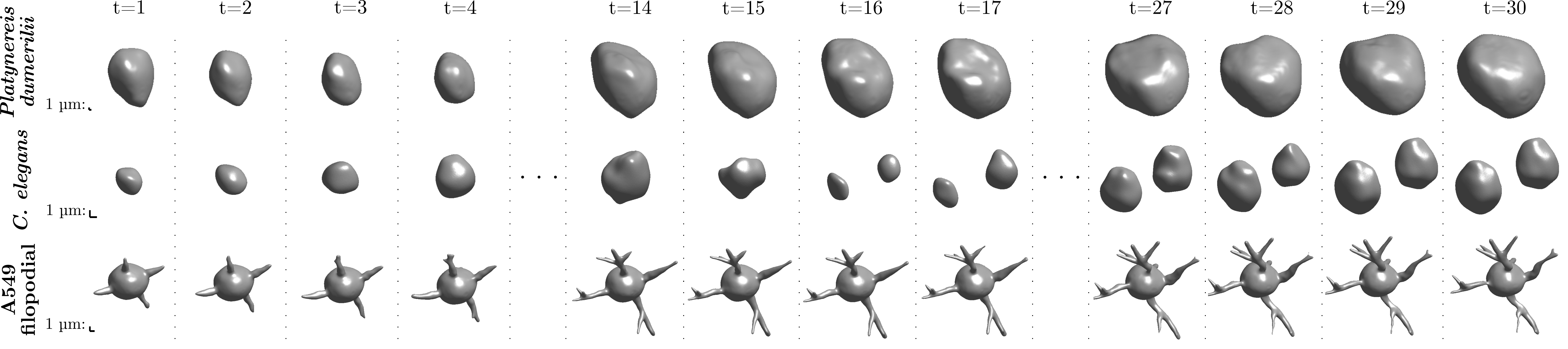}

\footnotesize{(b) Randomly generated synthetic cell shapes}

\vspace{-2mm}

\caption{\textbf{Visual comparison of real, reconstructed (a), and randomly generated synthetic cell shapes (b) in 3D+time.} \textmd{The proposed method is able to synthesize living cell shapes that accurately mimic processes such as cell growth in \textit{Platynereis dumerilii} cell (top), cell division in \textit{C. elegans} cells (middle), and growth and branching of filopodial protrusions in A549 lung cancer cells (bottom).}}
\label{fig:newsequences}
\end{figure*}

There have been several efforts to collect large community data sets of images and masks, such as the Broad Bioimage Benchmark Collection\footnote{\url{https://bbbc.broadinstitute.org}} or BioImage.IO\footnote{\url{https://bioimage.io/\#/?type=dataset}}. In case images are available but manual annotation is infeasible, masks generated by automatic algorithms are sometimes provided as a silver standard~\citep{ulmanreview,burgossvobodabook}. For example, this approach was taken for a variety of optical microscopy scenarios in the Cell Tracking Challenge\footnote{\url{http://celltrackingchallenge.net}}, where silver-standard corpora were used for training new segmentation methods~\citep{ctc:st_user:arbelleIsbi2019,ctc:st_user:loeffler2022}.

Alternatively, if neither images nor masks are available, both can be synthesized. Previously, fully synthetic annotated images were obtained with carefully hand-crafted models based on measurements in real images~\citep{2007lehmussola,2009svoboda,2012rajaram,2015malm,2016kovacheva,2016murphy,2016stegmaier,Svoboda:MitoGen2017,sorokin2018filogen}. Recently, deep learning methods, and in particular those using generative adversarial networks (GANs), have shown promising results for augmentation of existing training sets~\citep{2017osokin,2017goldsborough,2019bohland,2019bailo,2019baniukiewicz,2021kozlovsky} or synthesis of completely new data sets~\citep{2018fu,2019dunn}. The apparent strength of these deep learning-based approaches lies in their ability to learn information autonomously, usually the appearance of cells, and to create corresponding faithful new content.

The synthesis of new data pairs typically has two stages. The first stage consists of synthesizing a completely new shape (in 2D or 3D) or shape sequence (when time is included). In the second stage, these shapes are used to synthesize textured cell images that are visually similar to real microscopy images. GANs excel at this second stage owed in part to image-to-image translation methods such as \textit{pix2pix}~\citep{2017isola}, which have found widespread application in microscopy image synthesis. However, the first stage requires fundamentally different models and, most importantly, shape representations that can deal with the variation in cell shapes. As exemplified in Fig.~\ref{fig:newsequences}, cells exhibit a range of visually observable phenomena including global cell body changes such as cell growth, cell division, deformation in cell tissue, or projections of leading edge during cell motility, as well as localized body changes, such as blebbing (i.e., randomly growing and retracting blobs on the cell surface) or growing filopodia (i.e., highly motile thin protrusions budding off the main cell body). 

In this work, we synthesize living cell shapes in 3D+time, which we do from user-given examples such as segmentation masks of real cells. We propose to model the cell surface as the zero level-set of a continuous signed distance function (SDF) in 3D space and time. Following the DeepSDF model proposed by~\citep{park2019deepsdf}, we represent this SDF function in an implicit neural representation (INR). Such an INR consists of a multilayer perceptron (MLP) that we jointly optimize with a latent space using a large set of 3D cell shape time-lapse sequences. Once trained, owing to the continuous implicit representation, the optimized model can be used to synthesize completely new cell shape sequences at any given spatial or temporal resolution. In contrast to voxel-based or mesh-based cell shape representations with discretized sampling rates, our approach alleviates the limitations on spatial and temporal resolution and allows us to model and represent cell shapes at an arbitrary level of detail. By parametrizing the cell surface implicitly through differentiable and, therefore, trainable neural networks, the complexity of the resulting model is independent of the spatial and temporal resolution and instead scales with the complexity of the cell surface. Hence, in contrast to voxel-based or mesh-based representations, the resulting network has fixed memory requirements. 

We have previously proposed to represent living cell shapes using INRs~\citep{2022wiesner}. This work extends our previous work in several ways. First, we have adapted our model to be independent of spatial rotation of the cell. That is, we disentangle latent code and cell rotation so that two shapes that are rotated versions of each other will always be represented by the same latent code. This is a logical requirement in cell shape synthesis, where there is no canonical orientation, which allows us to learn more descriptive latent codes. Second, we have extended our data set with additional cell types so that we now optimize our model on three widely different cell shapes: \textit{Platynereis dumerilii} embryo cells, \textit{C. elegans} embryo cells, and A549 human lung carcinoma cells. Third, we have substantially extended our quantitative evaluation, and included an ablation study in which we evaluate the effect of rotation equivariance on the compactness of the learned latent space. We show how with these additions, we can generatively model different types of living cells with SO(3)-equivariant INRs and use the synthesized shapes as input to a GAN model to synthesize pairs of time-lapse images and segmentation masks.

\section{Related work}
We summarize existing explicit cell shape representation and synthesis methods and provide a brief introduction to related work in implicit neural representations. Moreover, we provide an introduction to equivariance and invariance in geometric deep learning models.

\subsection{Explicit representations}
Cell shape masks organized on pixel or voxel grids are the predominant standard for cell shape representation. The shape and resolution of the masks are chosen so that they match those of the original microscopy image, allowing easy correspondence matching and overlaying of the original image. While this has advantages, the memory requirements of such a representation grow quadratically (in 2D) or cubically (in 3D) with resolution. Hence, to represent fine details, such as a cell body with thin protrusions, memory requirements explode. This is also the case when the shape needs to be evolving~\citep{Svoboda:MitoGen2017}, in which case the denser grid makes it less likely for the shape boundary pixels to fall off the grid and to lose boundary localization precision. Voxel grids have been used in Cellular Potts Models~\citep{2005merks} (CPMs), which simulate time-resolved 3D cell shapes within a cell population~\citep{2005merks,2012swat,2014starruss,2020svoboda}. Deep learning models using GANs for cell shape synthesis also rely on grids~\citep{Wiesner:3dgan, 2018fu, 2019baniukiewicz}. These models are able to synthesize static cell shapes in 2D, pseudo-3D, and 3D. In the pseudo-3D approach, individual 2D slices are synthesized and subsequently composed into a 3D volume. One significant drawback of grid representations is that there are no topological guarantees, which might result in disconnected components. 

An alternative way to represent cell shapes is to represent them as a combination of (overlapping) spheres~\citep{dufour2005segmenting}. Several physics-oriented systems developed to simulate and study cell populations have opted to use this representation of the shape~\citep{van2015simulating,2018ghaffarizadeh}. Other lattice-free models use ellipsoids, in particular to represent cell nuclei~\citep{2019bohland,2019dunn,2019han}. A drawback of such representations is their limited ability to represent cells with protrusions, blebs, and other fine details. Polygonal meshes provide an alternative choice when detailed representation of 2D manifolds in 3D is desired. The approach is very well established for static shapes, but living cell shapes such as we consider in this paper are challenging to model with triangular meshes~\citep{2016li,sorokin2018filogen}, and often lead to intersecting faces and other mesh artifacts~\citep{2016li}. Alternatively, cells can be represented using spherical harmonics~\citep{ducroz2012:spharmat}.

\subsection{Implicit neural representations}
Implicit neural representations (INRs), also called \textit{coordinate networks} or \textit{neural fields}, have recently become a popular choice to represent signals in space and -- optionally -- time~\citep{xie2022neural}. INRs are based on the idea that a multilayer perceptron with coordinates as input can universally represent functions on a domain. However, in practice, multilayer perceptrons suffer from spectral bias, which means that they have difficulties representing high-frequency signals. Efforts to overcome this bias have focused on positional encoding of input coordinates~\citep{mildenhall2020nerf,tancik2020fourier} or the use of alternative activation functions~\citep{sitzmann2020implicit}. INRs can be used to represent any function in any space, which has led to a range of applications in medical imaging. For example, INRs can represent sinograms for CT reconstruction~\citep{sun2021coil}, MRI images obtained from k-space measurements~\citep{shen2022nerp}, deformation fields in image registration~\citep{wolterink2022implicit}, or 
outputs in image-to-image synthesis~\citep{chen2023local}. Of particular interest to the current work are INR representations of manifolds in space, typically via the implicit representation of an SDF. \citep{gropp2020implicit} showed how based on a limited number of training points, an INR can represent a continuous manifold in space. \citep{park2019deepsdf} demonstrated how an INR can be coupled with a latent space by conditioning the multilayer perceptron on a latent code, a feature that is critical to the current work. Recently, \citep{erkocc2023hyperdiffusion} proposed an approach for unconditional generative modeling using INRs. First, a set of MLPs is optimized to implicitly represent given data samples. Subsequently, a diffusion model is trained on the optimized weights of the MLPs to model the underlying distribution. The resulting diffusion model is then used to synthesize new MLP weights that represent an INR to generate new data samples in 3D and 3D+time.

\subsection{Equivariant and invariant shape learning}
Geometric deep learning is a learning paradigm in which neural networks are constructed under consideration of symmetry, i.e., by specifying groups of transformations to which the networks are supposed to be equivariant or invariant. A neural network is called equivariant if transforming its input results in the same transformation applied to its output and invariant if transforming the input has no effect on its output. For instance, when we use a neural network to classify images of cats and dogs, the neural network should be invariant to the rotations of the animals in the picture. Furthermore, when performing segmentation using a neural network, the neural network should output a rotated segmentation when the input image is rotated. In other words, the neural network should be equivariant to rotations. Finally, invariance and equivariance play a role in shape representation. For example, shapes do not intrinsically change when an object is rotated. Consequently, if we have two rotated instances of an otherwise identical shape, we want them to share the same representation that is independent of rotations. Such symmetry can be induced in neural networks in different ways, e.g., by imposing and solving linear constraints on the trainable parameters~\citep{FinziWelling2021}. In the context of conditional shape encoding and decoding, \citep{AtzmonNagano2022} constructed equivariant and invariant layers via so-called frame averaging. \citep{DengLitany2021} enforce rotational symmetry between latent code and shape by casting latent codes to Euclidean vectors for which rotation is well-defined. In this work, instead of coupling the latent code with shape orientation, we aim to explicitly de-couple it, achieving task-specific (approximate) independence of rotation.

\begin{figure*}[!ht]
\centering
\includegraphics[width=\textwidth]{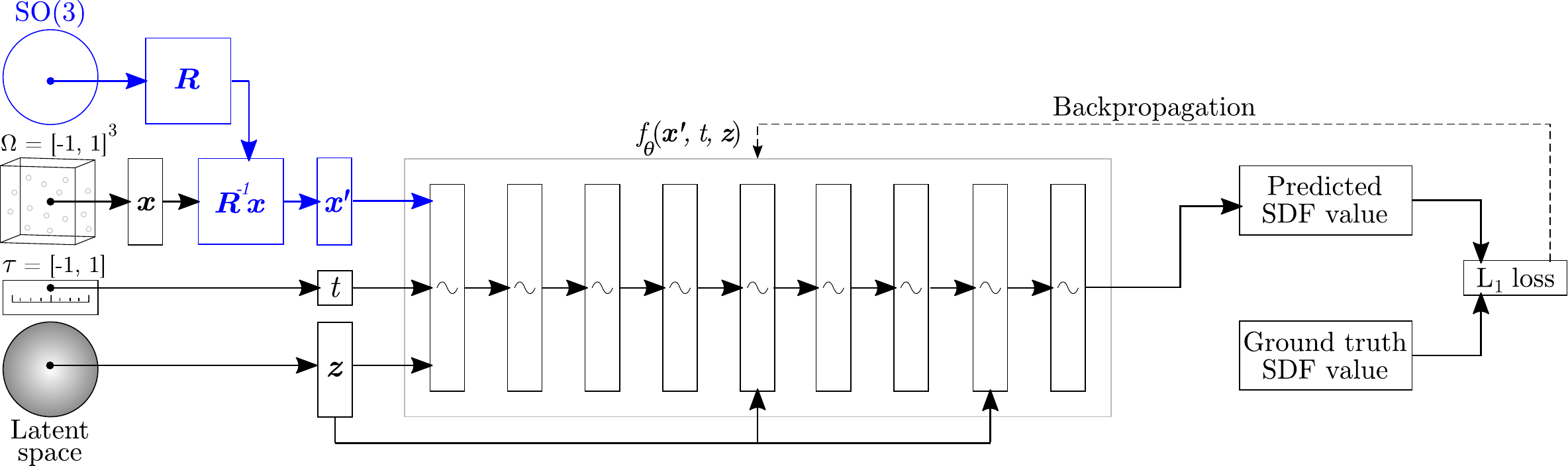}
\caption{\textbf{Conceptual diagram of the proposed network and its SO(3)-equivariant extension (in blue).} \textmd{The neural network $f_\theta$ is given a latent code $\bm{z}$ sampled from a~multivariate normal distribution, coordinates $\bm{x} = (x, y, z)$ from a~spatial domain $\mathrm{\Omega}$, and a temporal coordinate $t$ from temporal domain $\tau$. Moreover, the latent code $\bm{z}$ is given to the network not only at the input but also at its fifth and eight layer. The network is optimized to output the SDF values at given points, whereas the latent codes are jointly optimized to match a given normal distribution. The trained network is able to output SDF values based on a given latent code at any coordinate in the space-time domain. When given new latent codes from the latent space, the trained network is able to infer new spatio-temporal SDFs and thus produce new time-evolving shapes. The equivariant extension modifies the sampling procedure of the spatial coordinates and introduces a rotation matrix $\bm{R} \in \text{SO}(3)$. Rotating the spatial coordinates $\bm{x}$ by a rotation matrix $\bm{R}$ results in a new rotated spatial coordinates $\bm{x}'$. The rotation matrix $\bm{R}$ is optimized along with the network weights $\theta$ and the latent codes $\bm{z}$. During inference, the network reconstructs an SDF of a time-evolving cell shape according to a randomly sampled latent code $\bm{z}$ and given spatial coordinates $\bm{x}$, temporal coordinate $t$, and rotation matrix $\bm{R}$.}}
\label{fig:diagram_model}
\end{figure*}

\section{Method}
This work describes a method to model the surface of living cells in space and time. We represent the time-evolving cell surface implicitly as the zero level-set of a time-dependent continuous SDF that is parametrized by a neural network. We condition the neural network on learned latent code that describes the dynamics of the cell in space and time. By sampling new codes from the learned latent distribution and using these as input to the model, we can synthesize new and unseen time-evolving shapes. Furthermore, we disentangle the latent code and the rotation of a shape. This facilitates learning a more compact latent space as all rotations of a particular shape are represented by a single latent code, with rotation explicitly defined by a separate parameter. This results in a rotation equivariant implicit representation of time-evolving cell shapes. Figure~\ref{fig:diagram_model} shows a diagram of the network.

\subsection{Signed distance function}
We propose to represent the evolution of a cell surface as the zero-level set of its time-evolving SDF. The SDF is a continuous function that, for any point in space, gives the signed Euclidean distance to the nearest point on the cell surface. By convention, its sign is positive for points outside the shape and negative for points inside the shape. Here we also consider that the SDF of a living cell at a particular point in space evolves over time. More precisely, let $\mathrm{\Omega} = [-1, 1]^3$ be a spatial domain, $\tau = [-1, 1]$ a temporal domain, and  $\mathcal{M}_t$ be a 2D manifold embedded in $\mathrm{\Omega}$ at time $t \in \tau$. For any point $\bm{x} = (x, y, z)\in \mathrm{\Omega}$, the $\text{SDF}_{\mathcal{M}_t}:\mathrm{\Omega} \rightarrow \mathbb{R}$ is defined as
\begin{equation}
    \text{SDF}_{\mathcal{M}_{t}} (\bm{x}) = 
    \begin{cases}
    \min_{\bm{u} \in \mathcal{M}_{t}} ||\bm{x} - \bm{u}||_2, & \text{$\bm{x}$ outside $\mathcal{M}_{t}$} \\
    0, & \text{$\bm{x}$ belonging to $\mathcal{M}_{t}$} \\
    -\min_{\bm{u} \in \mathcal{M}_{t}} ||\bm{x} - \bm{u}||_2, & \text{$\bm{x}$ inside $\mathcal{M}_{t}$} 
    \end{cases}
\end{equation}
The zero-level set, and thus the surface of the cell at time $t$, is represented by all points where $\text{SDF}_{\mathcal{M}_t}(\cdot)=0$.

\subsection{Implicit neural representations}
Recent works have shown that the function $\text{SDF}_{\mathcal{M}_t}(\bm{x})$ can be approximated using a multi-layer perceptron (MLP) $f_\theta$ with trainable parameters $\theta$~\citep{sitzmann2020implicit,park2019deepsdf}. Such an MLP, called an INR, takes a coordinate vector $\bm{x}$ as input and provides an approximation of $\text{SDF}_{\mathcal{M}_t}(\bm{x})$ as output. We here propose to also condition the MLP on a time parameter $t \in \tau$ to provide an approximation of the time-evolving SDF of $\mathcal{M}_t$ for arbitrary $t \in \tau$. Hence, we regress the SDF values at a position at a certain time. In addition, the MLP can be conditioned on a latent space vector $\bm{z}$ drawn from a multivariate Gaussian distribution with a spherical covariance $\sigma^2 I$, where $I$ is the identity matrix. This latent code can be thought of as a low-dimensional encoding of a shape. By conditioning the MLP on a latent code, we are able to optimize a single model for a distribution of time-evolving shapes. 

Combining all these terms results in an MLP $f_\theta(\bm{x}, t, \bm{z})$ that approximates the time-evolving SDF of the manifold $\mathcal{M}_t$ for arbitrary $t \in \tau$, given latent space vector $\bm{z}$. Now, we describe how we optimize such a model for cell shape sequences $\mathcal{M}_t$ using an \textit{auto-decoder} strategy \citep{park2019deepsdf}.

\subsection{Network optimization}
\label{optimization}
We optimize the auto-decoder given a training set consisting of $N$ cell shape sequences $\{\mathcal{M}^i\}_{i = 1}^N$. For each cell shape sequence, reference values of its time-evolving SDF are known at a finite and discrete set of points in $\mathrm{\Omega}\times\tau$. An important aspect of the auto-decoder model is that not only the parameters $\theta$ of the MLP are optimized during training, but also the latent code $\bm{z}_i$ for each training sequence $\mathcal{M}^i$. The loss function therefore consists of two components. The first component is the reconstruction loss that computes the $L_1$ distance between reference SDF values and their approximation by the MLP,~i.e.,
\begin{equation}
\mathcal{L}_{rec}(f_\theta(\bm{x}, t, \bm{z}), \text{SDF}_{\mathcal{M}_t}(\bm{x})) = \|f_\theta(\bm{x}, t, \bm{z} ) - \text{SDF}_{\mathcal{M}_t}(\bm{x})\|_1,
\end{equation}
The second component is given by 
\begin{equation}
\mathcal{L}_{code}(\bm{z}) = \frac{1}{\sigma^2} \|\bm{z}\|^{2}_{2}.
\end{equation}
This term, with regularization constant $\frac{1}{\sigma^2}$, ensures that a compact latent space is learned and improves the speed of convergence~\citep{park2019deepsdf}. The parameter $\sigma^2$ in the regularization term $\mathcal{L}_{code}$ corresponds to the variance of the Gaussian distribution used for sampling the latent codes.
During training of the auto-decoder, we have access to a training set of $N$ cell shape sequences and thus the full loss function is
\begin{equation}\label{network}
\begin{array}{l}
\mathcal{L}(\theta, \{\bm{z}_j\}_{j=1}^N) = \\[1ex] \quad \mbox{\Large\( \mathbb{E}_{(\bm{x}, t)} \)}\left(\sum\limits_{i=1}^N \mathcal{L}_{rec}(f_\theta(\bm{x}, t, \bm{z}_i), \text{SDF}_{\mathcal{M}_t^i}(\bm{x})) +\mathcal{L}_{code}(\bm{z}_i)\right),
\end{array}
\end{equation}
where a latent code $\bm{z}_i$ is optimized for each shape $\{\mathcal{M}^i\}_{i = 1}^N$. After optimization, we obtain an MLP $f_\theta$ that is able to approximate the SDF of a shape $\mathcal{M}_{t}^i$, given latent vector $\bm{z}_i$, spatial coordinates $\bm{x}$, and time coordinate $t$, i.e.,
\begin{equation}
f_\theta(\bm{x}, t, \bm{z}_i) \approx \text{SDF}_{\mathcal{M}_{t}^i} (\bm{x}).
\end{equation}

\subsection{Network architecture}
\label{mod1_arch}
We represent the function $f_\theta(\bm{x}, t, \bm{z}_i)$ by an MLP, which can have any finite depth and width. Here, in all experiments, the MLP has 9 hidden layers, each containing 128 units. In the hidden layers, we use a \textit{sine} periodic activation function, which was shown to be able to better represent finely detailed surfaces of complex shapes compared to the commonly used rectified linear unit (ReLU)~\citep{sitzmann2020implicit, mildenhall2020nerf,2022wiesner}. The weights of layers using \textit{sine} activations are initialized with a~parameter $\omega_0$, which controls the angular frequency of the sine functions. This parameter directly affects the range of frequencies that the model is able to represent, where low values encourage low frequencies and smooth surfaces, and high values favor high frequencies and finely detailed surfaces. As proposed in~\citep{sitzmann2020implicit}, we initialize the weight matrices of shape ($c_{\textrm{out}} \times c_{\textrm{in}}$) in hidden layers by drawing from a uniform distribution
\begin{equation}
    \mathcal{U}\left(-\sqrt{\frac{6}{\omega_{0}^{2} \ c_{\textrm{in}}}}, \sqrt{\frac{6}{\omega_{0}^{2} \ c_{\textrm{in}}}} \ \right)
\end{equation}
while we draw the weights in the input layer from 
\begin{equation}
    \mathcal{U}\left(-\frac{1}{c_{\textrm{in}}}, \frac{1}{c_{\textrm{in}}} \ \right).
\end{equation}
We inject latent code vectors $\bm{z}_i$ in the first, fifth, and eighth layer of the network to improve reconstruction accuracy~\citep{park2019deepsdf}. Moreover, we concatenate the coordinates $\bm{x}$ and $t$ and inject them into all hidden layers. Preliminary experiments found this to be a requirement for convergence on long spatio-temporal sequences.

\subsection{Rotation equivariance} 
\label{sec:motivation_equivariant_model}
Thus far, we have described a model that can be optimized to jointly learn reconstructions and learn a latent space of cell shape sequences. However, during training, we assign each shape a latent code $\bm{z}$ and optimize the latent space only for compactness. Therefore, it might happen that a simple rotation of the same cell is described by $f_\theta(\bm{x}, t, \tilde{\bm{z}})$ where $\tilde{\bm{z}}\neq \bm{z}$ or that there is no latent code representing the rotated shape time series at all. In other words, even though a rotated shape time series is, in essence, the same shape time series, the latent code $\bm{z}$ does not represent the identity of the shape time series, and the model has to learn the rotated shape time series as well. To make sure that the rotated shape time series are also included in the model and to let $\bm{z}$ represent the identity of a shape time series, we propose the following \textbf{equivariant} model.

Let $\bm{R} \in \text{SO}(3)$ be a rotation matrix, where $\text{SO}(3)$ is the 3D rotation group representing all rotations about the origin in 3D Euclidean space. Every rotation matrix $\bm{R} \in \mathbb{R}^{3 \times 3}$ is orthogonal, i.e., its transpose is equal to its inverse that corresponds to a rotation in the opposite direction. Now, assume we rotate a shape described by the zero level set of $f_\theta(\bm{x}, t, \bm{z})$ via the rotation matrix $\bm{R}$. The resulting shape is described by $\bm{R}\{\bm{x} \mid f_\theta(\bm{x}, t, \bm{z}) = 0 \} = \{\bm{R}\bm{x} \mid f_\theta(\bm{x}, t, \bm{z}) = 0\}$ (where we let $\bm{R}$ act on the whole set by abuse of notation). By substituting $\bm{x'} = \bm{R} \bm{x}$, we then see that the rotated shape is described by $\{\bm{x'} \mid f_\theta(\bm{R}^T\bm{x'}, t, \bm{z}) = 0\}$, the zero level set of the function $f_\theta(\bm{R}^T \cdot, t, \bm{z})$. Hence, to represent rotated shape time series, we only need to apply a (transposed) rotation matrix to $\bm{x}$ in $f_\theta(\bm{x}, t, \bm{z})$. 

As seen in Fig. \ref{fig:diagram_model}, we achieve (approximate) equivariance by adding a rotation matrix $\bm{R}$ to our model that transforms our SDF $f_\theta(\bm{x}, t, \bm{z})$ to the SDF corresponding to the rotated shape time series $f_\theta(\bm{R}^T\bm{x}, t, \bm{z})$. We optimize this model similarly to the \textit{non}-equivariant model, with some important differences. First, we assign a rotation matrix $\bm{R}$ to each training time series and optimize this rotation matrix during training. We parametrize $\bm{R}$ using rotation angles $\alpha_i$, $\beta_i$, $\gamma_i$ around axes $x$, $y$, and $z$, respectively. Using $\bm{R}_{\varphi_i}$ as the rotation matrix induced by the rotation angles $\varphi_i = (\alpha_i, \beta_i, \gamma_i)$, the loss function we obtain is:
\begin{equation}\label{network2}
\begin{array}{l}
\mathcal{L}(\theta, \{\bm{z}_j\}_{j=1}^N, \{(\alpha_j, \beta_j, \gamma_j)\}_{j=1}^N) = \\[1ex] \ \mbox{\Large\( \mathbb{E}_{(\bm{x}, t)} \)}\Biggl(\sum\limits_{i=1}^N \mathcal{L}_{rec}(f_\theta(\bm{R}_{\varphi_i}^T \bm{x}, t, \bm{z}_i), \text{SDF}_{\mathcal{M}_t^i}(\bm{x})) + \mathcal{L}_{code}(\bm{z}_i) \Biggl).
\end{array}
\end{equation}
We optimize angles $\alpha_i$, $\beta_i$, $\gamma_i$ along with latent codes $\bm{z}_i$ and network weights $\theta$. The angles $\alpha_i, \beta_i, \gamma_i$ are initialized from $\mathcal{N}(0,\frac{\pi^2}{64})$ as suggested in~\citep{bepler2019explicitly}. Note that we do not assume a canonical orientation for cell shapes, and angles can be different for each cell sequence.

\subsection{Proving equivariance}
\label{equivariance}
So far, we have not discussed the rationale behind calling this new model equivariant. To see this, we note that the latent code and rotation matrix $\bm{R}$ belonging to a new shape time series $\mathcal{M}_t$ are found by minimizing the following loss function over $\varphi = (\alpha, \beta, \gamma)$ and $\bm{z}$:
\begin{equation}
    \mbox{\Large\( \mathbb{E}_{(\bm{x}, t)} \)}\Biggl(\mathcal{L}_{rec}(f_\theta(\bm{R}_{\varphi}^T \bm{x}, t, \bm{z}), \text{SDF}_{\mathcal{M}_t}(\bm{x})) + \mathcal{L}_{code}(\bm{z}) \Biggl).
\end{equation}
or, more precisely, its estimate using $M$ SDF tuples $\{(\bm{x}_i, t_i, s_i)\}_{i=1}^M$ of $\mathcal{M}_t$ with $s_i = \text{SDF}_{\mathcal{M}_{t_i}}(\bm{x}_i)$:
\begin{equation}
    \frac{1}{M}\sum_{i=1}^M\Biggl(\mathcal{L}_{rec}(f_\theta(\bm{R}_{\varphi}^T \bm{x}_i, t_i, \bm{z}), s_i) + \mathcal{L}_{code}(\bm{z}) \Biggl)
    \label{eq:opt_prob_fin_case}
\end{equation}
When minimizing the latter loss function over $(\bm{R}_\varphi, \bm{z})$, we have the following equivariance property:

\begin{theorem}
Let the reconstruction of a shape time series $\mathcal{M}_t$ be obtained by minimizing Equation \ref{eq:opt_prob_fin_case} over $(\bm{R}_\varphi, \bm{z})$. Denote by $\mathcal{M}_{t}^{\tilde{\bm{R}}}$ the shape time series $\mathcal{M}_t$ rotated by applying the rotation matrix $\tilde{\bm{R}}$. Assume $(\bm{R}, \bm{z})$ is a solution to the optimization problem for $\mathcal{M}_t$ and let the data points for $\mathcal{M}_{t}^{\tilde{\bm{R}}}$ be given by \[\{(y_i, t_i, s_i)\}_{i=1}^M = \{(\tilde{\bm{R}}x_i, t_i, s_i)\}_{i=1}^M.\] Then, $(\tilde{\bm{R}}\bm{R}, \bm{z})$ is a solution to the optimization problem for $\mathcal{M}_{t}^{\tilde{\bm{R}}}$.
\label{thm:equivariance_fin_case}
\end{theorem}

A proof of the above theorem and a proof of an infinite sample version of the theorem can be found in the supplementary material. Theorem \ref{thm:equivariance_fin_case} tells us that when rotating the time series of a shape and the spatial part of the used data $\{(\bm{x}_i, t_i, s_i)\}_{i=1}^M$ by a rotation matrix $\Tilde{\bm{R}}$, a good reconstruction of the rotated time series can be found by changing the obtained rotation matrix in an equivariant way while leaving the latent code $\bm{z}$ invariant. In this way, in our equivariant model, the latent code $\bm{z}$ truly represents the identity of the shape time series while $\bm{R}$ captures the rotational part.

\section{Data}
To demonstrate the ability of our proposed method to model different phenomena occurring during the cell cycle, we selected three diverse 3D time-lapse biomedical data sets. First, \textit{Platynereis dumerilii} cells exhibit rapid non-rigid cell shape deformations over time. Second, \textit{C. elegans} cells were selected to demonstrate growth of the embryo and clear divisions of mother cells into their daughters. Third, A549 carcinoma cells feature growing and branching protrusions on a blebbing cell main body. Each cell type thus exhibits distinct shape features that we want to model using the proposed method. The 3D+time data sets were acquired in fluorescence microscopy, and the microscopy images were complemented with full segmentation masks. The segmentation masks for \textit{Platynereis dumerilii} and \textit{C. elegans} cells were produced using automatic image analysis algorithms. The A549 carcinoma cells are computer generated with the segmentation masks produced jointly with the synthetic microscopy images. Here, we describe each data set in more detail and discuss the data preparation procedure used to produce suitable SDF data for training the models.

\subsection{Platynereis dumerilii embryo cells}
\textit{Platynereis dumerilii} is a sea worm that lives in tropical coastal areas and reaches a length from 2 to 4 cm when fully grown. It is commonly used in biological evolution studies as a model organism. The fluorescently-stained nuclei of a developing \textit{Platynereis dumerilii} embryo were acquired in its early stage of development. The acquisition was done using a SiMView light sheet microscope with double illumination and double detection objectives~\citep{tomer2012quantitative}. Specifically, using the illumination objectives Olympus XLFLUOR 4$\times$/340/0.28, and the detection objectives Nikon CFI75 LWD 16$\times$/0.8 W. The acquisition was done with a time step of 90 seconds for 300 time points, making the overall experiment time 7 hours and 30 minutes. The spatial resolution of the images is $700\!\times\!660\!\times\!113$ voxels, with a voxel size of $0.406\times0.406\times2.031$~{\textmu m} in the $x$, $y$, and $z$ axis, respectively.

\subsection{C. elegans embryo cells}
\textit{Caenorhabditis elegans} is a transparent worm living in temperate soil environments and reaching approximately 1 mm in length. Its molecular and evolutionary biology was extensively studied~\citep{brenner1974genetics} and, to this day, it is a widely used model organism in biological studies. The nuclei of \textit{C. elegans} embryo cell population were fluorescently stained in its early stage of development~\citep{2008murray} and acquired using a Zeiss LSM 510 Meta confocal laser scanning microscope with Plan-Apochromat 63$\times$/1.4 (oil) objective lens. The images were acquired with a 60 second time step over 250 time points with the overall experiment time being 4 hours and 10 minutes. The spatial resolution of the acquisition is $708\times512\times35$ voxels with voxel size of $0.09\!\times\!0.09\!\times\!1.0$~{\textmu m}. This data set is freely available from the website of the Cell Tracking Challenge~\citep{ulman2017objective}.

\subsection{A549 human lung carcinoma cells}
The A549 lung carcinoma cell line was cultivated from samples of cancerous human lung tissue. It is commonly used as a model in cancer studies and in the testing and development of drug therapies. This is a synthetic data set of simulated GFP-actin-stained A549 lung cancer cells embedded in a Matrigel matrix~\citep{sorokin2018filogen}. Both the membrane of a cell and its growing and branching filopodial protrusions are fluorescently stained. The acquisition process is simulated using a virtual Zeiss Axiovert 200M inverted fluorescence microscope with a Yokogawa CSU-10 confocal unit and Zeiss 40$\times$/1.30 (oil) objective. The time step is 20 seconds with 30 time points resulting in the overall experiment duration of 10 minutes. The spatial resolution is $300\!\times\!300\!\times\!300$ voxels with a voxel size of $0.125\!\times\!0.125\!\times\!0.125$~{\textmu m}. This data set was generated using the CytoPacq web-interface~\citep{wiesner2019cytopacq}. We simulated 33 time-lapse sequences, where each sequence captures one growing A549 filopodial cell.

\subsection{Data preprocessing}
\label{preprocessing}
To prepare suitable training data sets, we processed the segmentation masks of the \textit{Platynereis dumerilii} cells, the \textit{C. elegans} cells, and the A549 filopodial cells. We partitioned these masks into voxel volumes, where each voxel volume contains a single cell shape. Moreover, for each cell, we prepared these voxel volumes for the first 30 time points from its inception. Cell shapes were centered at each time point according to their centroid and aligned according to their principal axes. For each cell type, we prepared 33 time-lapse sequences with 30 time points. Note that for the \textit{C. elegans} cells, the prepared voxel volumes contain two daughter cells in the second half of the time-lapse sequence due to the mitosis of a selected mother cell. In this case, each daughter cell was aligned separately. 

We subsequently precomputed SDFs for each time point in these time-lapse sequences, i.e., for each sequence we computed 30 three-dimensional SDFs. Each time point was then represented by $256\!\times\!256\!\times\!256$ discrete SDF point samples. One time-evolving cell is thus represented by $30\times256^3$ samples, defining 30 time points of its 3D shape. These point samples constitute the training data sets, $\mathcal{D}_{Plat}^{SDF}$ of \textit{Platynereis dumerilii} cells, $\mathcal{D}_{Cele}^{SDF}$ of \textit{C. elegans} cells, and $\mathcal{D}_{Filo}^{SDF}$ of A549 filopodial cells. 3D renders of time-evolving shapes from the training sets are shown in Fig.~\ref{fig:newsequences}a. All data preparation and visualization algorithms were implemented\footnote{The source code, pre-trained networks, and data sets are available online at \url{https://cbia.fi.muni.cz/research/simulations/implicit_shapes} and on GitHub at \url{https://github.com/MIAGroupUT/IMPLICIT-CELL-SURFACES}.} in Matlab R2021a.

\section{Experiments and Results}
We present a series of experiments with the proposed auto-decoder and evaluate the results both quantitatively and qualitatively. Specifically, we investigate the reconstruction of cell sequences, synthesis of new shapes with randomly sampled latent codes, temporal interpolation between consecutive time points, and compactness of the learned latent spaces. Finally, we demonstrate how synthetic shapes can be used as input to an image-to-image model that synthesizes corresponding microscopy images.

\vspace{0.2cm}

\noindent\textbf{Training parameters} We trained separate models on $\mathcal{D}_{Plat}^{SDF}$, $\mathcal{D}_{Cele}^{SDF}$, and $\mathcal{D}_{Filo}^{SDF}$. All models were trained for 2000 epochs with a batch size of 5. More precisely, we select a time-evolving cell 5 times with repetition and subsequently select an individual time point from each selected time-evolving cell. The weights were initialized with $\omega_0$ set to 30, following the initialization scheme proposed in ~\citep{sitzmann2020implicit}, and optimized using the Adam optimizer~\citep{kingma2014adam} with a learning rate $10^{-4}$, which was reduced every 350 epochs by a factor of $0.5$. The latent codes were initialized randomly from $\mathcal{N}(0,0.01^2)$, as suggested in~\citep{park2019deepsdf}, and their dimensionality was set to 64. The given hyperparameters were used in all experiments unless stated otherwise. The networks and the respective training and inference procedures were implemented in Python using PyTorch~\citep{paszke2019pytorch} and PyTorch3D~\citep{johnson2020accelerating}.

\vspace{0.2cm}

\begin{figure}[!t]
\centering
\includegraphics[width=0.85\columnwidth]{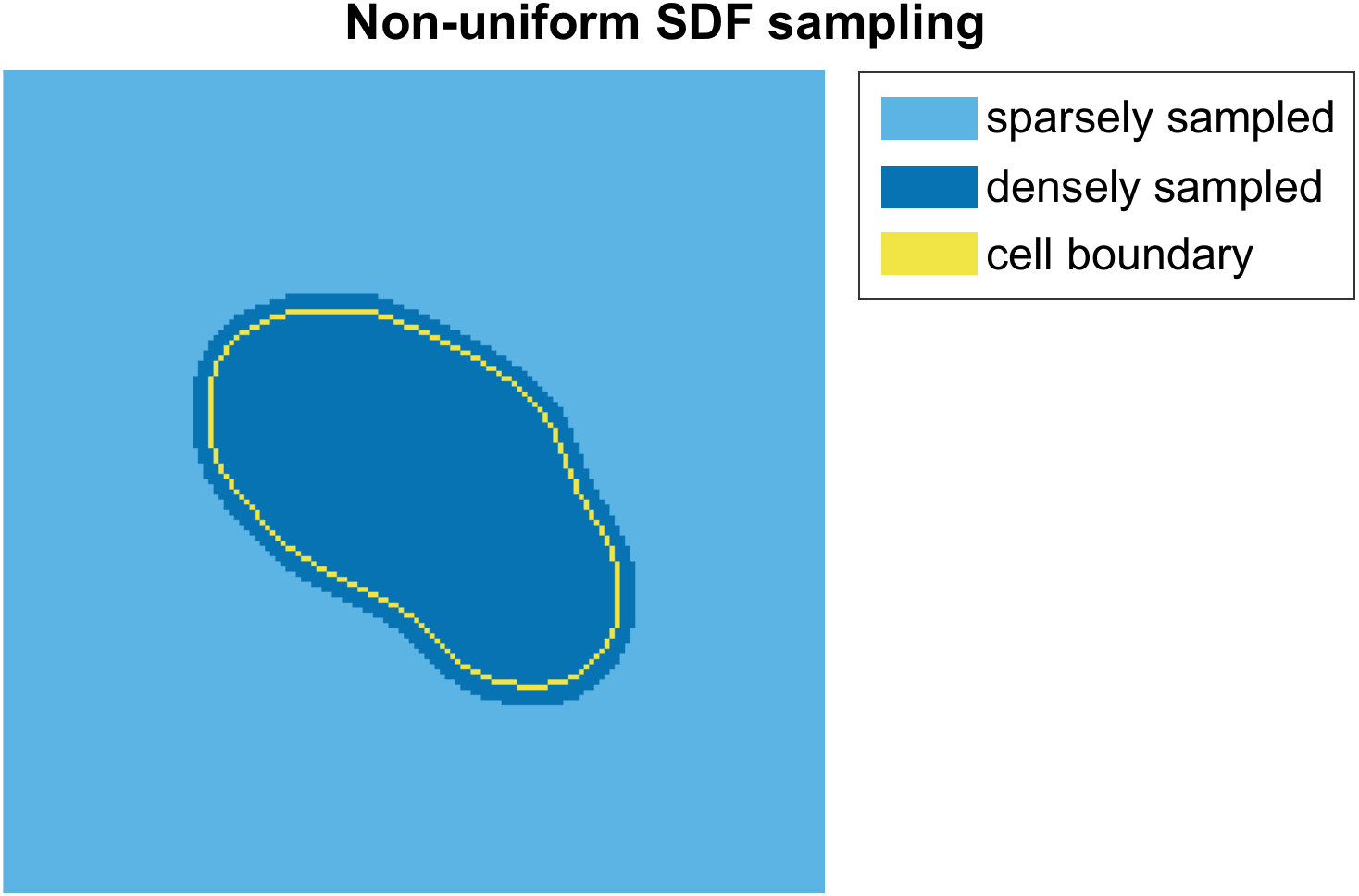}
\caption{\textbf{Non-uniform SDF sampling.} \textmd{The figure illustrates the non-uniform SDF sampling process used during training.
As the cell shape occupies only a fraction of the space, non-uniform sampling allows to reduce the number of point samples by focusing on the points that contain the most important information.}}
\label{figrev:sdfsampl}
\end{figure}

\noindent\textbf{SDF sampling} We randomly and non-uniformly sampled 1,000,000 SDF point samples per time point at each training epoch. The sampling procedure is illustrated in Fig.~\ref{figrev:sdfsampl}. To determine the sampling parameters, we analyzed the training data sets. As stated in Sec.~\ref{preprocessing}, the training data sets consist of $256^3$ discrete SDF point samples per time point. The number of point samples (mean $\pm$ standard deviation) representing the cell surface and its interior are 260,903 $\pm$ 186,188 in $\mathcal{D}_{Plat}^{SDF}$, 515,533 $\pm$ 260,459 in $\mathcal{D}_{Cele}^{SDF}$ (considering both daughter cells), and 294,265 $\pm$ 41,406 in $\mathcal{D}_{Filo}^{SDF}$. The values were obtained by counting the points $x$ where $\textrm{SDF}(x) <= 0$ at each time point over all sequences in the given data set.
As the cell shape occupies only a fraction of the considered 3D space, we sampled from the training data sets non-uniformly, as suggested in~\citep{park2019deepsdf}. Specifically, 70\% of a training batch is composed of points with distance to the cell surface less or equal to 0.6~[\textmu\text{m}] (including the points with negative distance representing the cell interior) to ensure that the cell boundary and thus the zero level-set is well represented. Conversely, the remaining points with a distance greater than 0.6 [\textmu\text{m}] form the remaining 30\% of a batch, as they represent only the empty space around the cell.
The non-uniform sampling allowed us to reduce the overall number of point samples and thus decrease the time and memory needed to train the models. Additionally, it also made the training optimization procedure focus more on the important points, as the $L_1$ distance in the reconstruction loss $\mathcal{L}_{rec}$ is computed against these point samples.

\vspace{0.2cm}

\noindent\textbf{Inference parameters} Because the trained model is continuous, it can be used to generate point clouds, meshes, or voxel volumes~\citep{park2019deepsdf}. In this work, we use binary volumes of size $256\!\!\times\!\!256\!\!\times\!\!256$ voxels for quantitative evaluation and meshes obtained using marching cubes~\citep{lorensen1987marching} for visualization. Each voxel volume and mesh represents a shape of a time-evolving cell at a given time point. With a trained model, generating an SDF of a time-evolving shape with $30\!\times\!256\!\times\!256\!\times\!256$ point samples took approximately 15 seconds on an NVIDIA A100 and required 3 GB of GPU memory. The stated memory consumption represents a forward pass through the MLP. It is composed of the network parameters and corresponding computational graph, along with the network inputs (i.e., a latent code and a grid of spatial and temporal coordinates) and the network outputs representing the inferred SDF values at given spatio-temporal coordinates. The proposed MLP has 168.5k parameters and is thus an order of magnitude smaller than convolutional generative models such as DCGAN~\citep{radford2015unsupervised} with 3.5M parameters. For inference, we divide the coordinate grid of $30\times256^3$ points into subsets of $1\!\times\!32^3$ points. These subsets are given to the MLP, and the network outputs are subsequently concatenated to form the resulting SDF of a time-evolving shape. The size of the grid subset can be set depending on the available GPU memory and has no impact on the quality of resulting SDF as points are processed individually.

\vspace{0.2cm}

\noindent\textbf{Evaluation parameters}
For the reconstruction of cell shapes, we compute the Dice similarity coefficient (DSC). To compare distributions of cells, we compute descriptive statistics (i.e., minimum, maximum, mean, median value, and interquartile range) of shape descriptors, namely surface~{[\textmu$\text{m}^2$]}, volume~{[\textmu$\text{m}^3$]}, and sphericity~\citep{costa2000shape}, where sphericity ranges from 0 to 1, with $1$ representing an ideal sphere. Please note that we accounted for the cell division present in $\mathcal{D}_{Cele}^{SDF}$ by splitting each respective voxel volume sequence into two sequences in order to evaluate the shape of each daughter cell separately. Each evaluated sequence of voxel volumes thus represents a single time-evolving cell shape.

\begin{figure*}[!t]
\centering
\includegraphics[width=\textwidth]{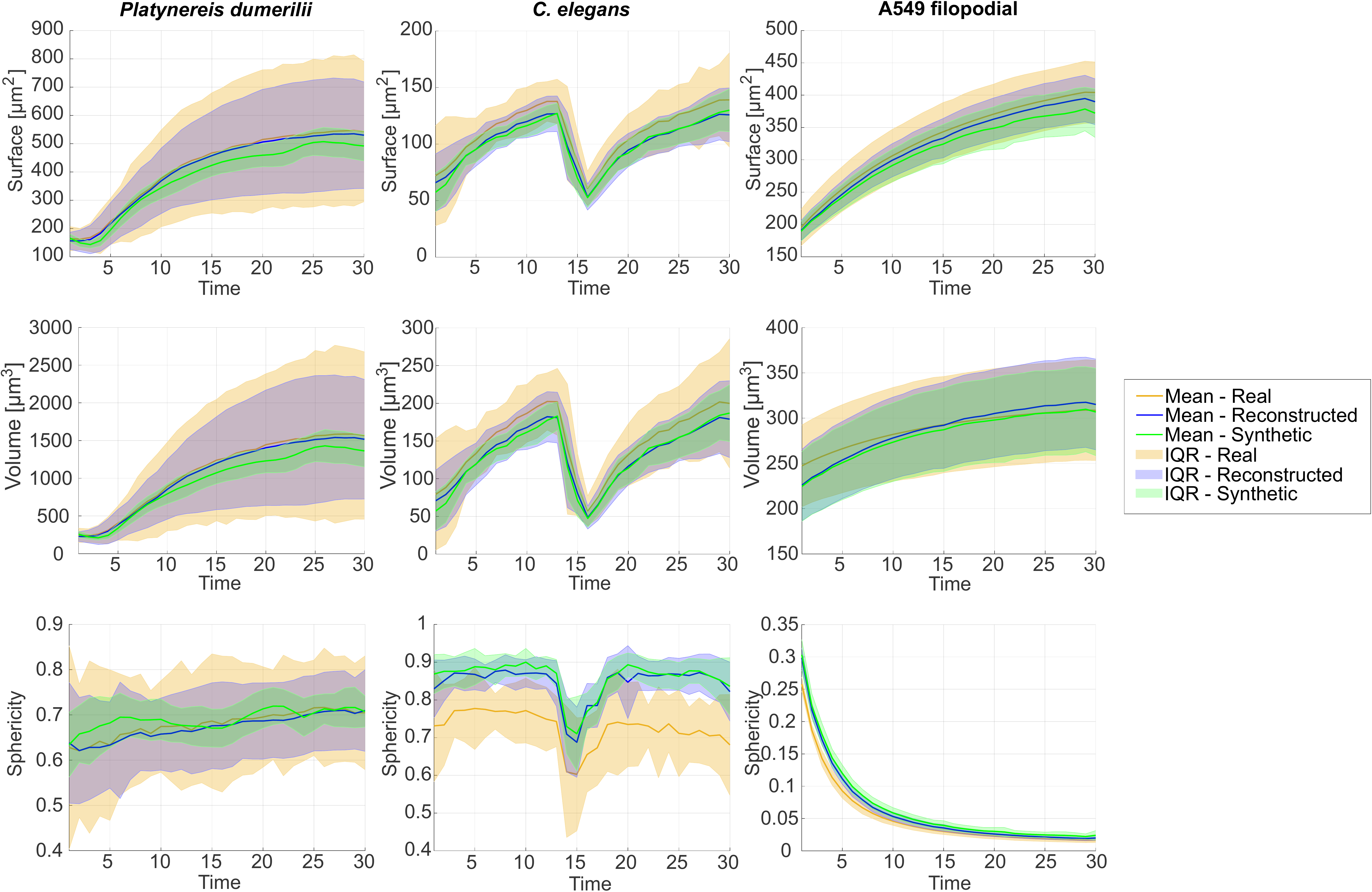}
\caption{\textbf{Quantitative evaluation of reconstructed cell shapes produced using the proposed equivariant model.} \textmd{The plots show  descriptive statistics of shape descriptors, i.e., surface ~{[\textmu$\text{m}^2$]}, volume ~{[\textmu$\text{m}^3$]}, and sphericity, at each time point in the time-lapse sequence for a given cell type (30 timepoints in total). From the left, the columns represent \textit{Platynereis dumerilii} cells, \textit{C. elegans} cells, and A549 filopodial cells. Real shapes from the training sets are denoted as ``Real'', shapes reconstructed by the model from the optimized latent codes are denoted as ``Reconstructed'', and shapes generated by random sampling are denoted as ``Synthetic''. Each plot shows the mean and interquartile range (IQR) of the respective shape descriptor, with the values computed at each given time point over all time-lapse sequences (33 in total). The closer are the respective colored plotted values together, the more similar the shapes are, where yellow stands for shapes from the real set, and blue for the reconstructed shapes.}}\label{fig:reconstructions}
\end{figure*}

\begin{figure*}[!t]
\centering
\includegraphics[width=\textwidth]{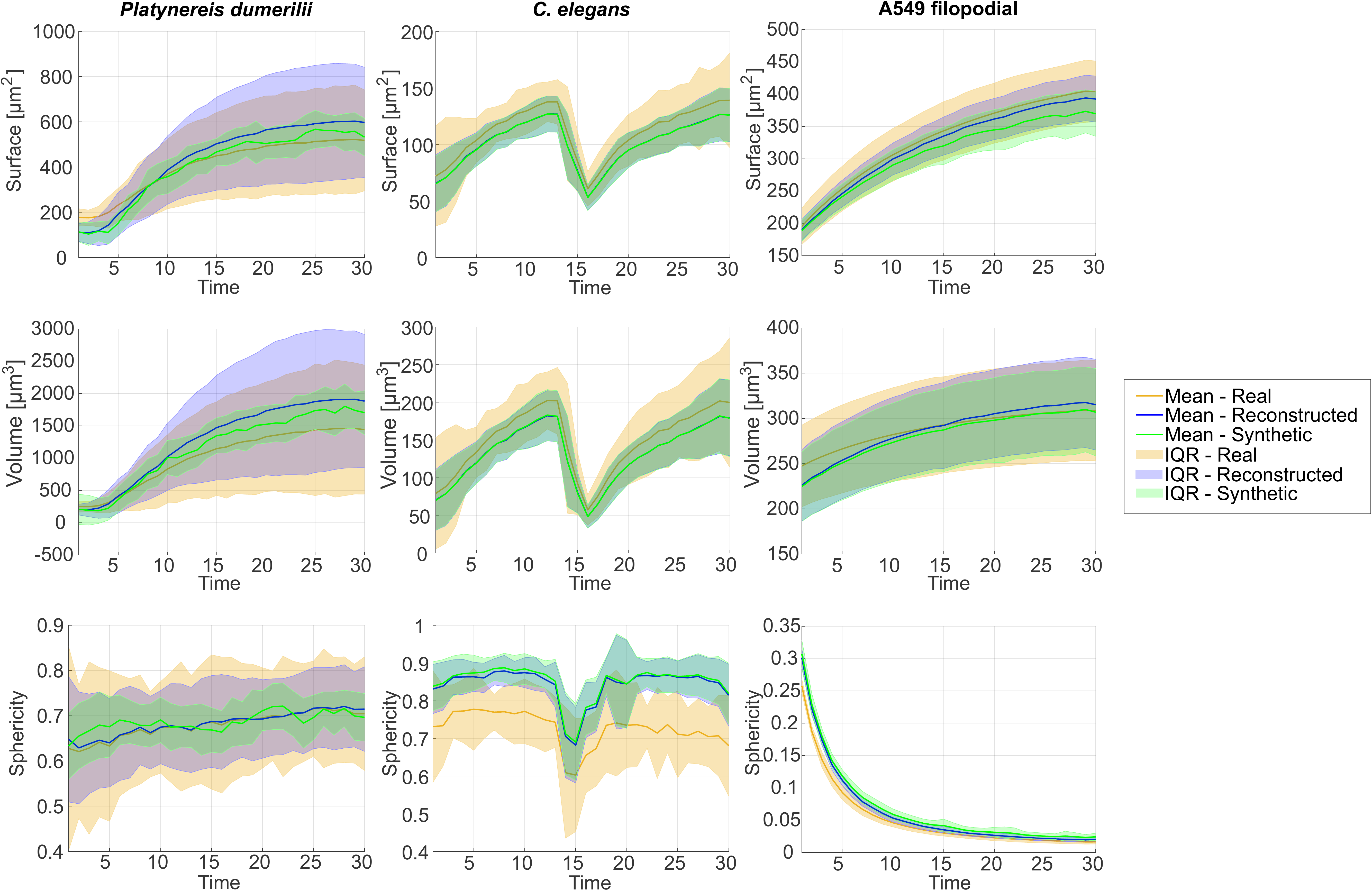}
\caption{\textbf{Quantitative evaluation of reconstructed cell shapes produced using the \textit{non}-equivariant model.} \textmd{The plots show  descriptive statistics of shape descriptors, i.e., surface ~{[\textmu$\text{m}^2$]}, volume ~{[\textmu$\text{m}^3$]}, and sphericity, at each time point in the time-lapse sequence for a given cell type (30 timepoints in total). From the left, the columns represent \textit{Platynereis dumerilii} cells, \textit{C. elegans} cells, and A549 filopodial cells. Real shapes from the training sets are denoted as ``Real'', shapes reconstructed by the model from the optimized latent codes are denoted as ``Reconstructed'', and shapes generated by random sampling are denoted as ``Synthetic''. Each plot shows the mean and interquartile range (IQR) of the respective shape descriptor, with the values computed at each given time point over all time-lapse sequences (33 in total). The closer are the respective colored plotted values together, the more similar the shapes are, where yellow stands for shapes from the real set, and blue for the reconstructed shapes.}}\label{fig:reconstructions_noeq}
\end{figure*}

\subsection{Reconstruction and synthesis of time-evolving shapes}
\label{sec:reconstruction}
In this experiment, we evaluated the ability of the models to reconstruct a cell shape sequence given its learned latent code $\bm{z}$ and to synthesize a new cell shape sequence given randomly sampled latent code. A separate model was optimized on $\mathcal{D}_{Plat}^{SDF}$, $\mathcal{D}_{Cele}^{SDF}$, and $\mathcal{D}_{Filo}^{SDF}$ to learn an implicit representation of \textit{Platynereis dumerilii} cells, \textit{C. elegans} cells, and A549 filopodial cells, respectively.

Fig.~\ref{fig:reconstructions} and Fig.~\ref{fig:reconstructions_noeq} show descriptive statistics for cell surface, volume, and sphericity over all time-lapse sequences between real, reconstructed, and randomly generated synthetic shapes produced by the equivariant model and the \textit{non}-equivariant model, respectively. The plots for reconstructed \textit{Platynereis dumerilii} shapes show an excellent overlap of the descriptive statistics. The synthetic shapes fit within the interquartile range of the real data set while keeping similar mean values and generally exhibit lower variability compared to the real ones while retaining the mean volume and surface as the cells grow over time.

The graphs for \textit{C. elegans} show that the descriptive statistics are similar, with a moderate shift toward lower values for the surface of reconstructed shapes. Conversely, we can observe that the sphericity is higher. Note the mitosis that the cells undergo at time point 15. We can observe from the plots that shapes get smaller and less round before growing again in the second part of the time-lapse sequence, and the reconstructed shapes reproduce this behavior very closely. The synthetic shapes similarly show a moderate shift toward the lower surface and volume while exhibiting higher sphericity. The models were able to accurately reproduce the cell division occurring in the middle of the time-lapse sequence. The visualization of the mitotic division can be seen in Fig.~\ref{fig:newsequences}a.

The reconstructed filopodial cells also exhibit excellent similarity. The reconstruction accurately retains the cell volume. We can observe moderately higher sphericity and lower surface area. This is due to the growing and branching filopodial protrusions that are thin and have sharp edges. The reconstructed filopodial cells exhibit slightly rounder edges and thus lose a moderate amount of their surface area. The synthetic shapes exhibit excellent overlap with respect to the cell volume and, similarly to the reconstruction, a moderately lower surface and higher sphericity.

The results show that the models are able to accurately reproduce cell growth of \textit{Platynereis dumerilii}, mitosis of \textit{C. elegans}, and growing and branching protrusions of A549 lung cancer cells. Randomly generated synthetic shapes exhibit reasonably similar shape features and high visual similarity to the real ones for all three cell lines, as shown by the visual comparison in Fig.~\ref{fig:newsequences}b.

Additionally, Table~\ref{tab:ks_rec} lists the \textit{p}-values of the two-sample Kolmogorov Smirnov (KS) test computed on the shape descriptors of the real and reconstructed shapes produced using the equivariant model. The KS test retained the null hypothesis ($p > 0.05$) that the shape descriptors are from the same distribution at $5\%$ significance level for all tests except for the sphericity of the synthetic \textit{C. elegans} and A549 human carcinoma cells, reflecting our observations.

\begin{table}[!t]
\caption{\textbf{\textit{p}-values of the two-sample Kolmogorov-Smirnov test computed on the shape descriptors of the real and reconstructed shapes produced using the proposed equivariant model.}}\label{tab:ks_rec}
\centering
\resizebox{0.36\textwidth}{!}{\begin{tabular}{|c|c|c|c|}
\hline
    \multirow{2}{4em}{\textbf{Cell type}} & \multicolumn{3}{|c|}{\textbf{\textit{p}-value}} \\ &  \multicolumn{1}{c}{Surface} & \multicolumn{1}{c}{Volume} & \multicolumn{1}{c|}{Sphericity} \\
\hline
    \textit{Platynereis dumerilii} & $0.634$ & $0.252$ & $0.387$ \\
\hline
    \textit{C. elegans} & $0.077$  & $0.472$ & $<0.001$ \\
\hline
    A549 filopodial &  $0.064$ & $0.230$ & $<0.001$ \\
\hline
\end{tabular}}
\end{table}

\vspace{2mm}

\noindent\textbf{Random sampling of latent codes} To generate new synthetic cell shapes, we randomly sampled new latent codes within the optimized latent space. For \textit{Platynereis dumerilii} and \textit{C. elegans}, we randomly sampled 33 new latent codes $\bm{z}$ from $\mathcal{N}(0,0.001)$. For filopodial cells, noise vectors sampled from $\mathcal{N}(0,0.0005)$ were added to the optimized latent codes obtained after training.

The parameters of the normal distribution were obtained by analyzing the latent codes of the models trained on $\mathcal{D}_{Plat}^{SDF}$ $\mathcal{D}_{Cele}^{SDF}$, and $\mathcal{D}_{Filo}^{SDF}$ data set. For the first two data sets, the models yielded latent codes with standard deviations of $0.0004$ and $0.0008$, respectively. We rounded these numbers up to $0.001$ to ensure good coverage of the learned latent spaces and used this value to define the normal distribution for sampling new codes. The model trained on $\mathcal{D}_{Filo}^{SDF}$ yielded latent codes with a standard deviation of $0.0007$. In this case, randomly sampled latent codes resulted in cells with incorrectly placed protrusions, regardless of the chosen normal distribution. Each cell in the filopodial data set has distinctly growing protrusions, and the shared similarity is only the main body of the cells. By random sampling in the latent space, the model correctly inferred the common features, in this case, the main cell body. However, the model was not able to learn a sufficiently generalized spatial and temporal representation of the protrusions, which vary in placement, branching, and growth direction between all given cells. We experimentally determined that sampling new latent codes within the proximity of already learned latent codes yields correct shapes that exhibit reasonable variability.

\begin{figure*}[!t]
\centering
    
    \captionsetup[subfigure]{justification=centering} 
    \begin{subfigure}[b]{0.395\textwidth}
        \centering
        \includegraphics[width=\textwidth]{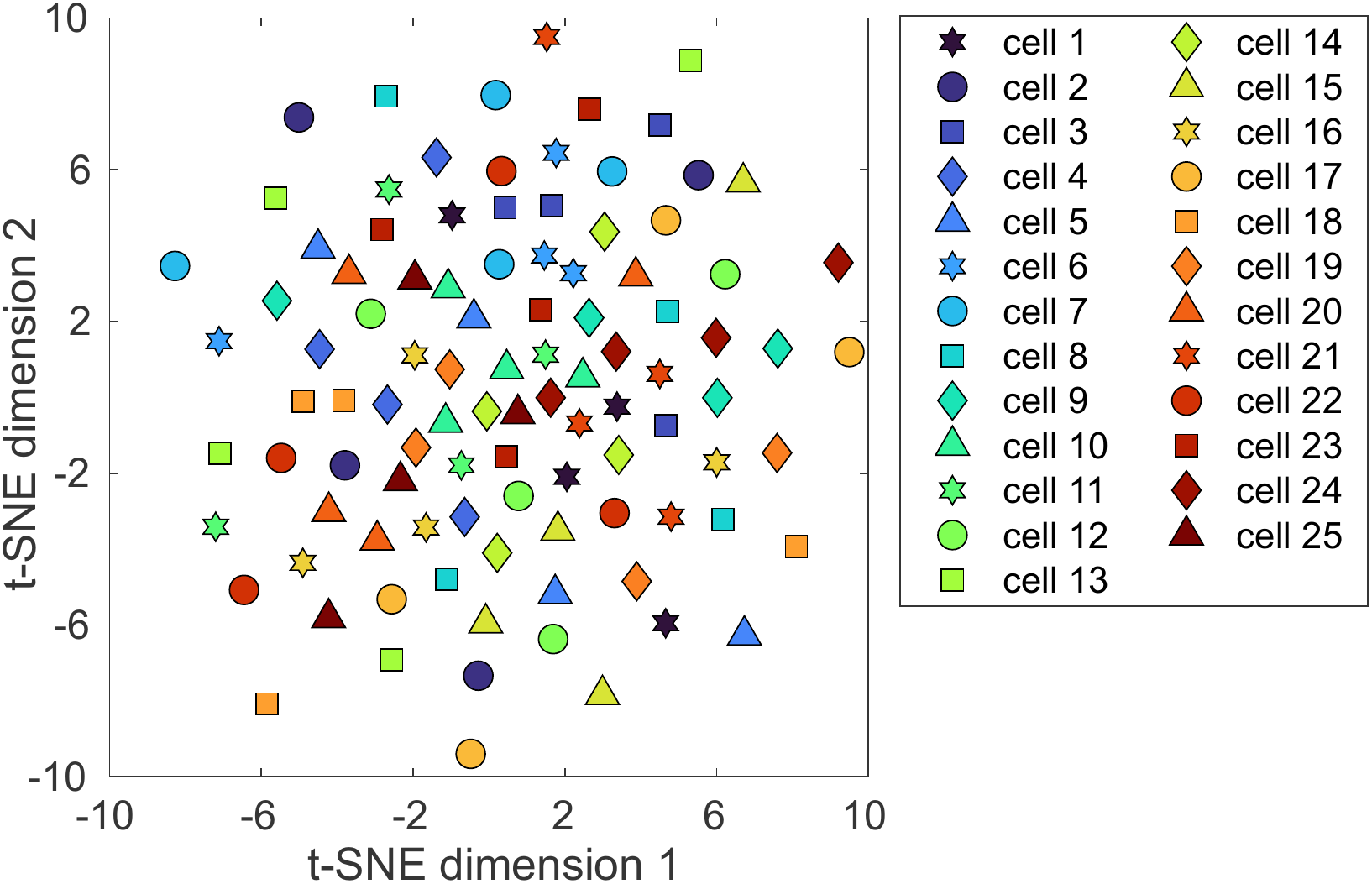}
        \caption{\textit{Non}-equivariant model}
        \label{figrev:latent_plat:noeq}
    \end{subfigure}
    \hspace{1cm}
    \begin{subfigure}[b]{0.395\textwidth}
        \centering
        \includegraphics[width=\textwidth]{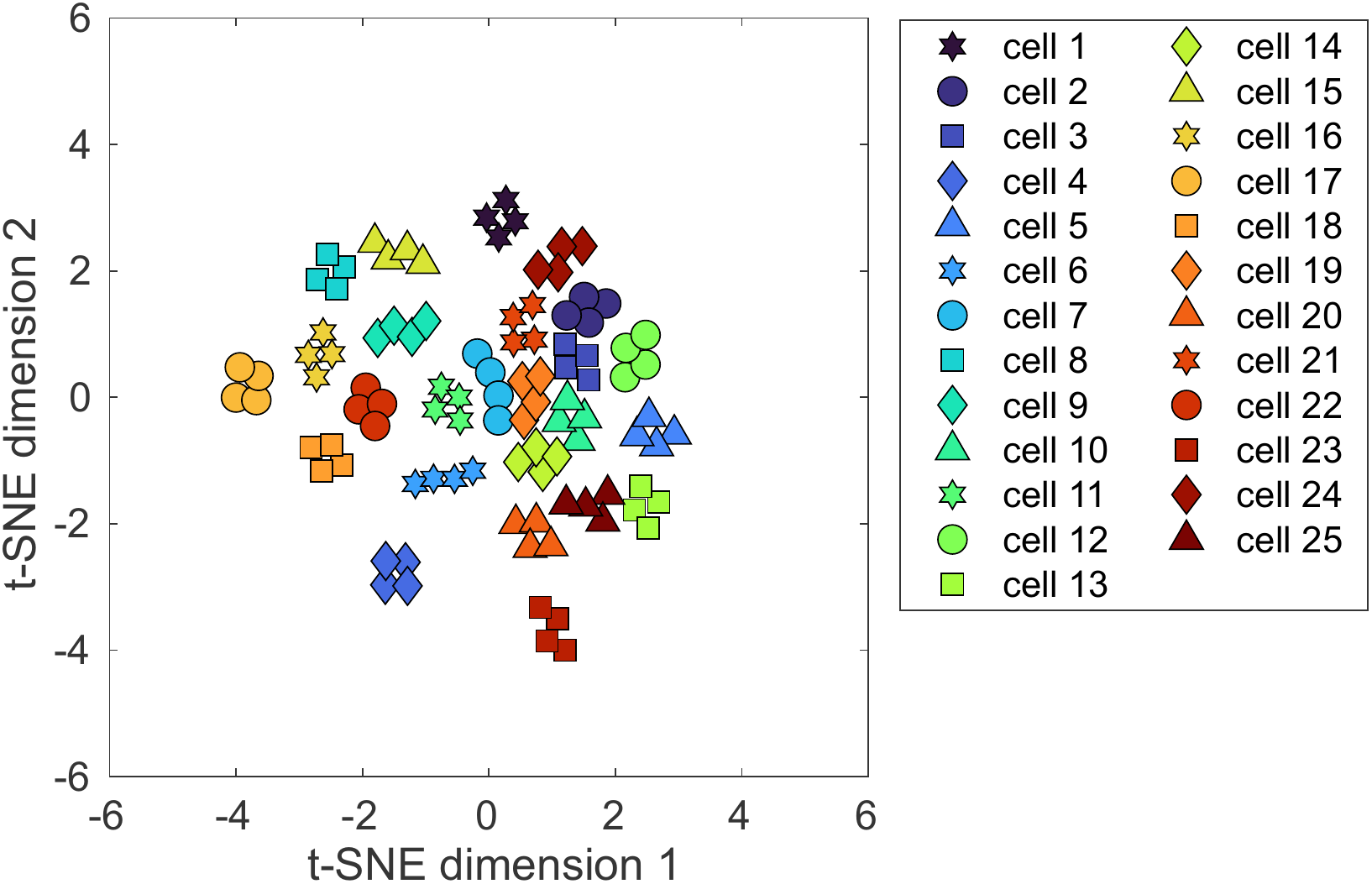}
        \caption{Equivariant model}
        \label{figrev:latent_plat:eq}
    \end{subfigure}

\vspace{-2mm}

\caption{\textbf{Latent space of the \textit{non}-equivariant model (a) and latent space of the equivariant model (b).} \textmd{The figure shows a comparison of the latent spaces on a low dimensional representation computed using t-SNE. There are 25 time-evolving cell shapes that have each been randomly rotated four times.}
}
\label{figrev:latent_plat}
\end{figure*}

\begin{figure*}[!t]
\centering

    \captionsetup[subfigure]{justification=centering} 
    \begin{subfigure}[b]{0.37\textwidth}
        \centering
        \includegraphics[width=\textwidth]{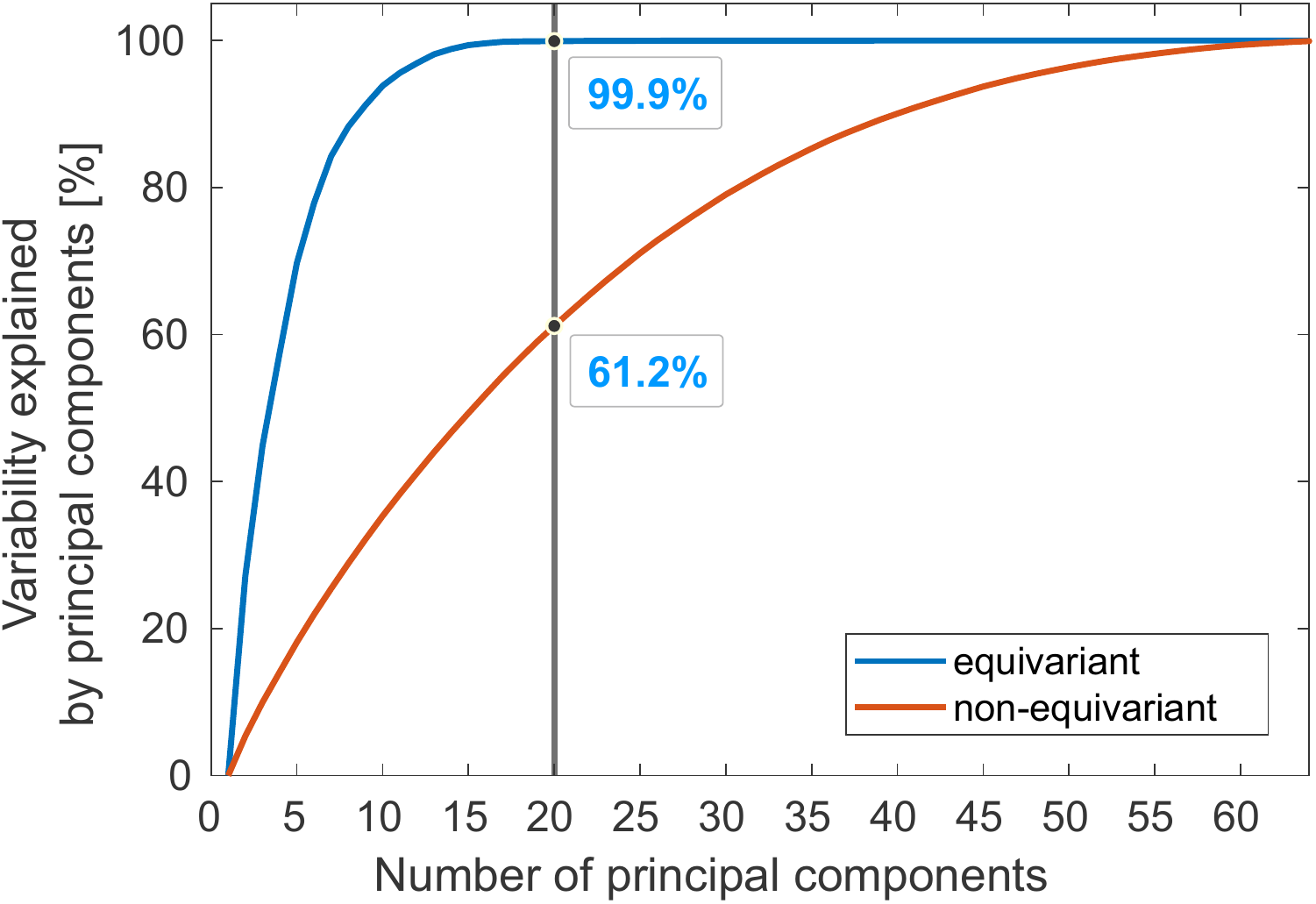}
        \caption{Principal component analysis}
        \label{figrev:latent_plat2:pca}
    \end{subfigure}
    \hspace{1cm}
    \begin{subfigure}[b]{0.4\textwidth}
        \centering
        \includegraphics[width=\textwidth]{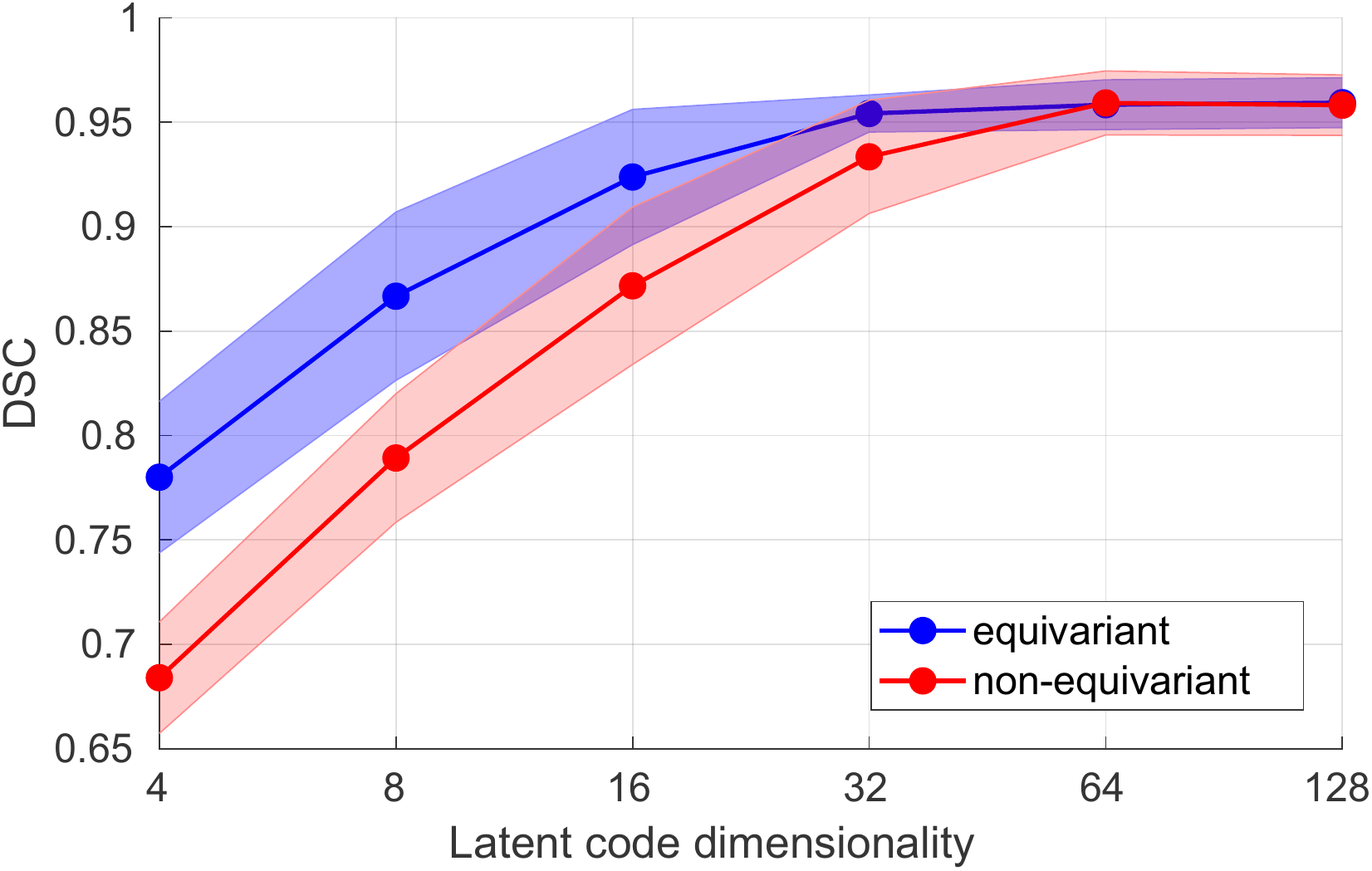}
        \caption{Dice similarity coefficient (DSC) with respect to the latent code dimensionality}
        \label{figrev:latent_plat2:dsc}
    \end{subfigure}

\vspace{-2mm}
    
\caption{\textbf{Principal component analysis (PCA) of the latent codes learned by the equivariant and the \textit{non}-equivariant model (a) and a plot of Dice similarity coefficient of real shapes and shapes reconstructed using each model with respect to the latent code dimensionality (b).} \textmd{The PCA plot shows the amount of information present in the latent codes, whereas the DSC plot shows how the latent code dimensionality affects the reconstruction results of the models.}
}
\label{figrev:latent_plat2}
\end{figure*}

\begin{table*}[t!]
\caption{\textbf{Dice similarity coefficient (DSC) (mean ± standard deviation) of shapes reconstructed using the equivariant and the \textit{non}-equivariant model with respect to latent code dimensionality.} \textmd{Results were computed over reconstructed shapes at time points 30, where the cells are fully grown, over 33 time-lapse sequences. The values show the similarity of the reconstructed cell shapes to the shapes from the training data set. An ideal identical shape would have DSC equal to $1$. Cases in which a specific model yielded better results are highlighted in bold.}}\label{tab:latent_plat2}
\centering
\resizebox{0.8\textwidth}{!}{\begin{tabular}{|r||c|c|c|c|c|c|}

\hline
    \textbf{Latent dimensionality} & \textbf{4} & \textbf{8} & \textbf{16} & \textbf{32} & \textbf{64} & \textbf{128} \\
\hhline{|=======|}
    \textbf{DSC -- equivariant model} & \textbf{0.780$\pm$0.036} & \textbf{0.867$\pm$0.40} & \textbf{0.924$\pm$0.32} & \textbf{0.954$\pm$0.009} & 0.958$\pm$0.012 & \textbf{0.959$\pm$0.112} \\
\hline
    \textbf{DSC -- \textit{non}-equivariant model} & 0.684$\pm$0.27 & 0.789$\pm$ 0.031 & 0.872$\pm$0.038 & 0.934$\pm$0.027 & \textbf{0.959$\pm$0.015} & 0.958$\pm$0.015 \\
\hline

\end{tabular}}
\end{table*}

\begin{figure*}[!t]
\centering
\captionsetup[subfigure]{justification=centering} 
    
    \begin{subfigure}[b]{0.33\textwidth}
        \centering
        \includegraphics[width=\textwidth]{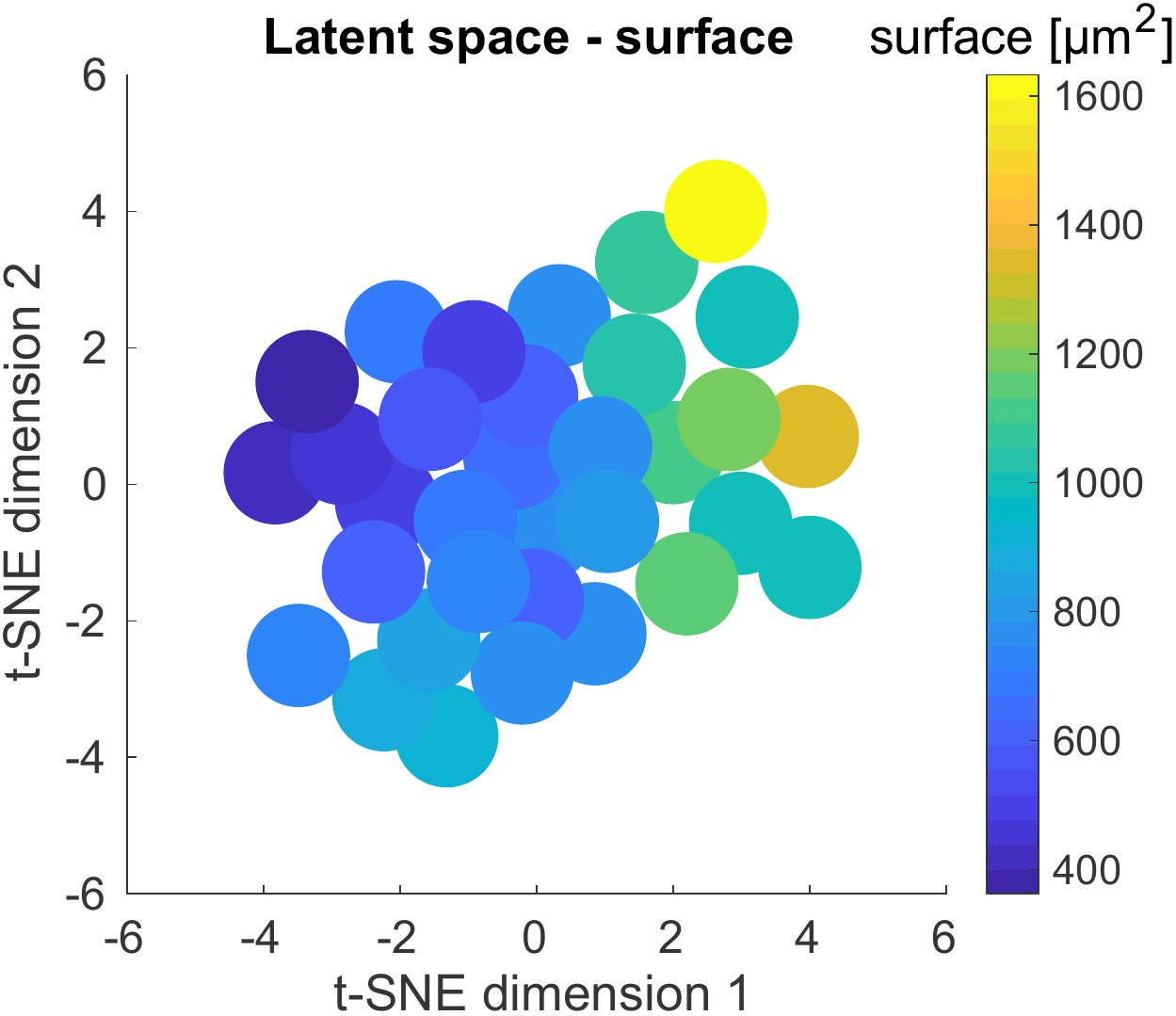}
        \caption{}
        \label{figrev:latent_explore:surf}
    \end{subfigure}
    \hfill
    \begin{subfigure}[b]{0.33\textwidth}
        \centering
        \includegraphics[width=\textwidth]{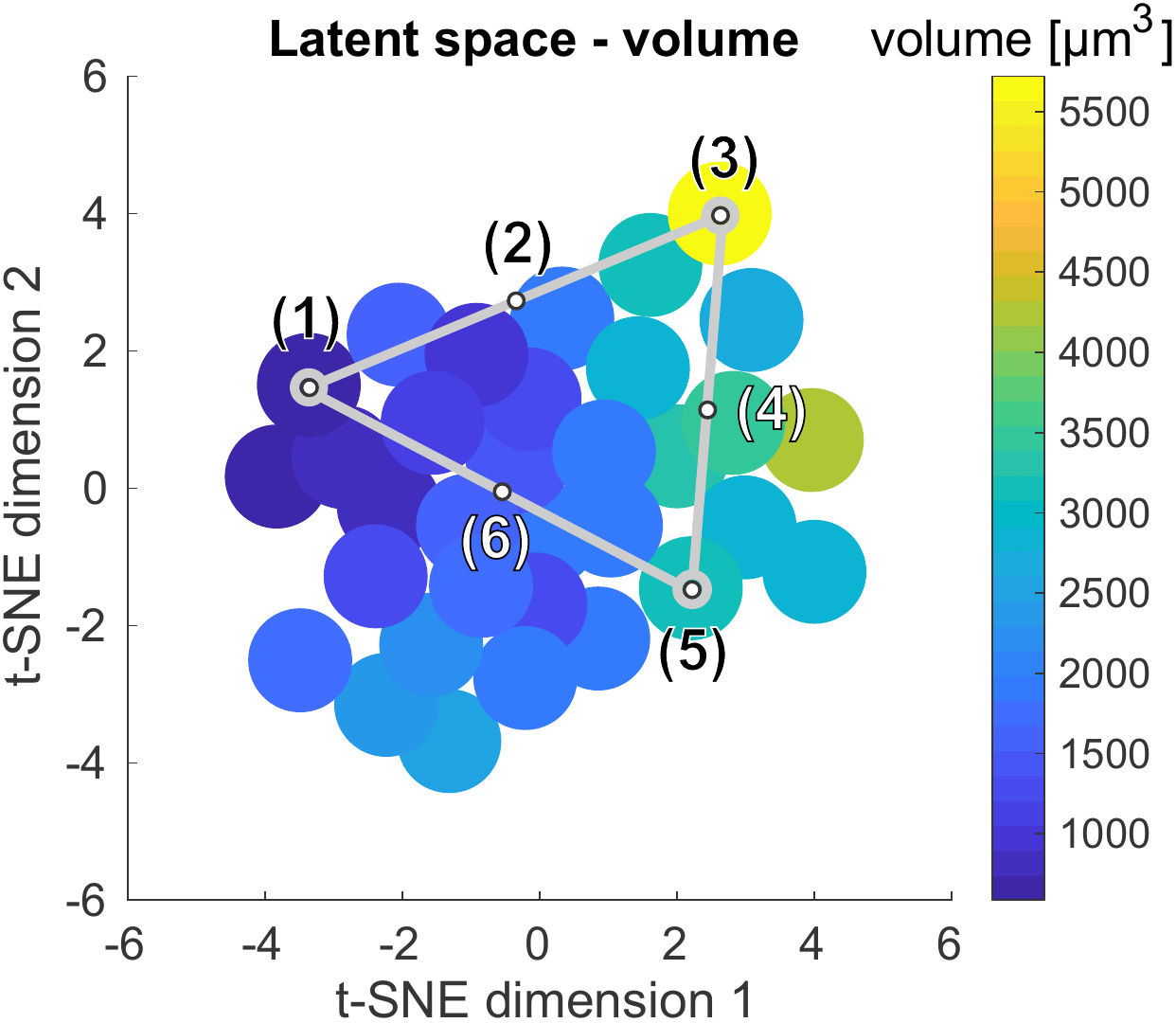}
        \caption{}
        \label{figrev:latent_explore:vol}
    \end{subfigure}
    \hfill
    \begin{subfigure}[b]{0.33\textwidth}
        \centering
        \includegraphics[width=\textwidth]{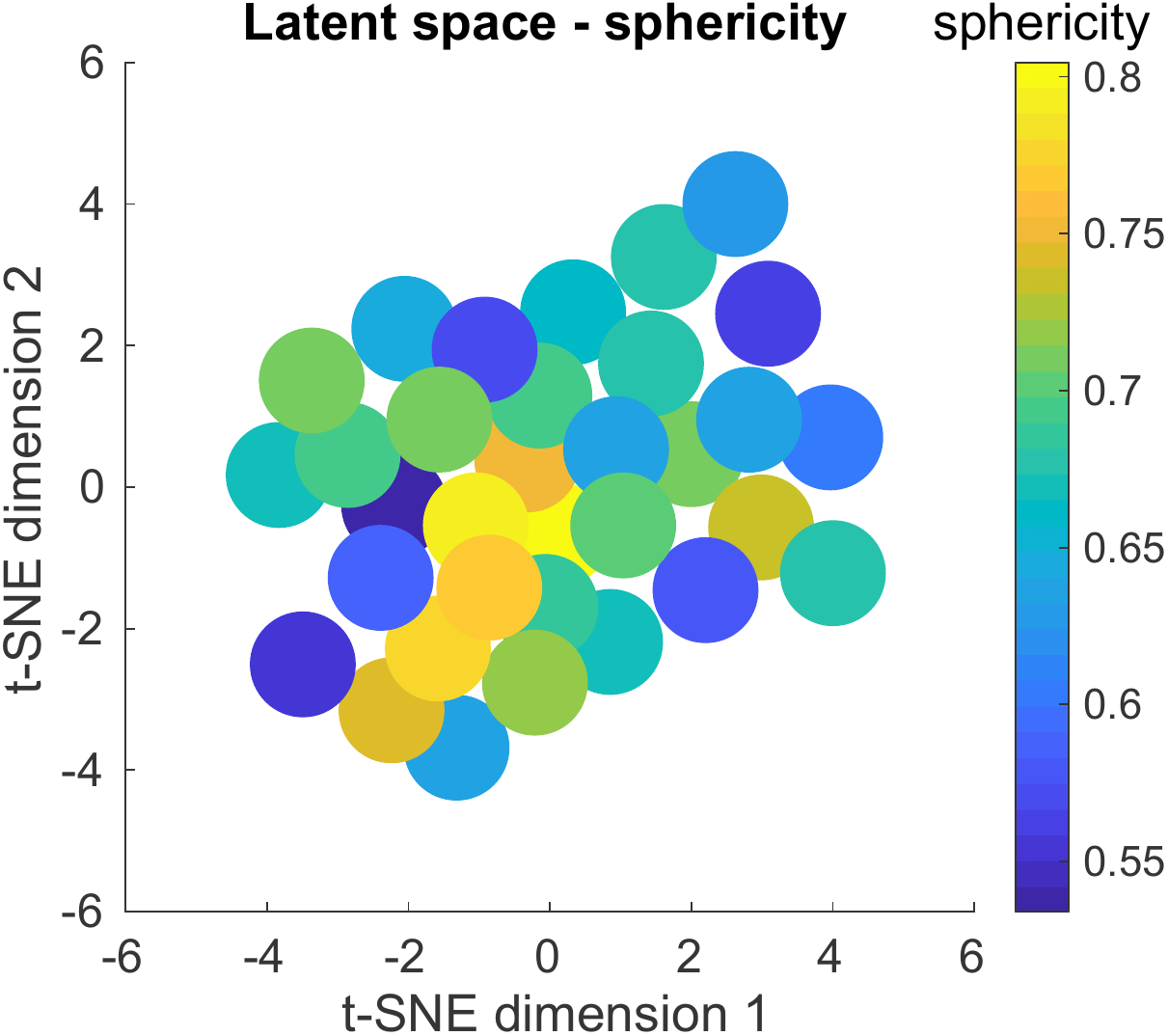}
        \caption{}
        \label{figrev:latent_explore:sph}
    \end{subfigure}

    \begin{subfigure}[b]{\textwidth}
        \centering
        \includegraphics[width=\textwidth]{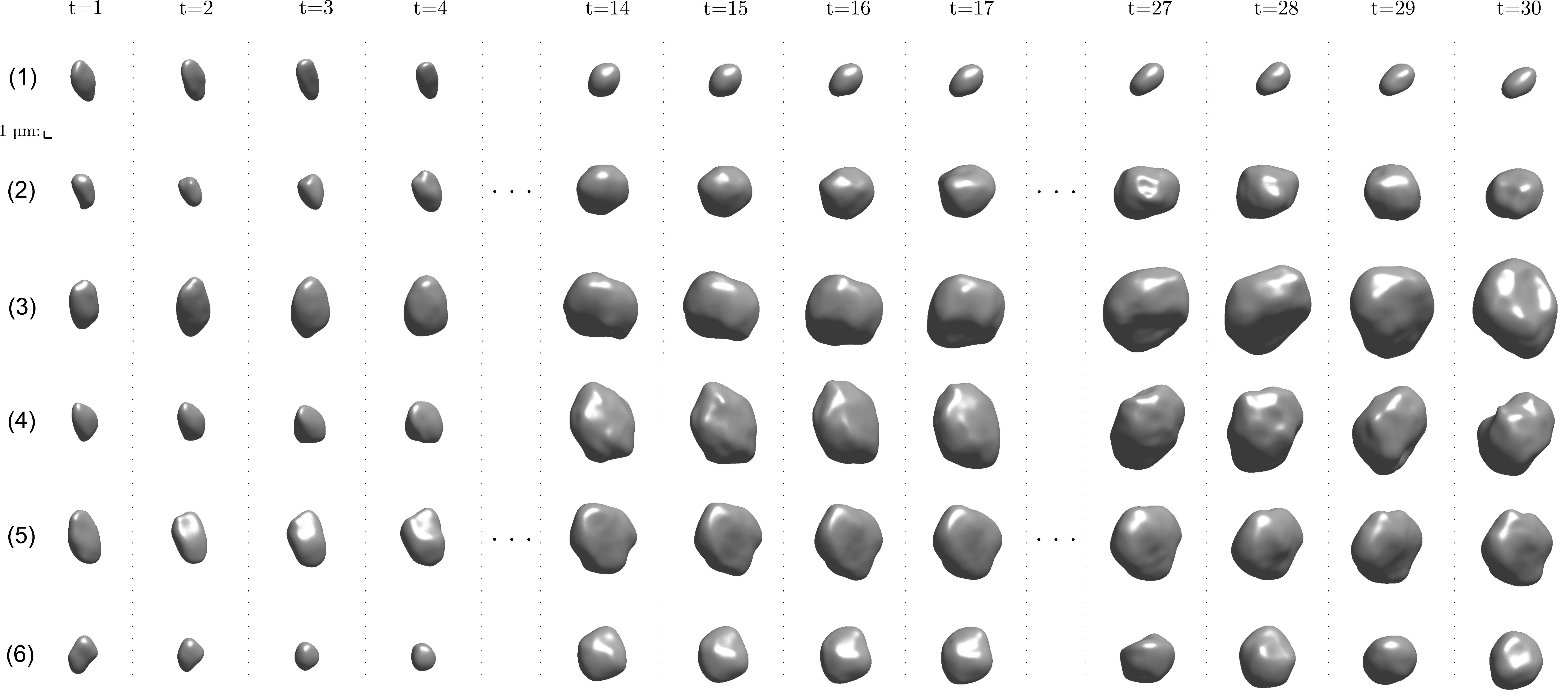}
        \caption{Time-evolving shapes with mean cell volume determined by the latent code}
        \label{figrev:latent_explore:sdf}
    \end{subfigure}

\vspace{-2mm}
     
\caption{\textbf{Latent space exploration.} \textmd{The figure shows a low dimensional representation of the latent space learned by the equivariant model trained on $\mathcal{D}_{Plat}^{SDF}$ and a visualization of the reconstructed cell shapes sampled from selected locations in the latent space. The low dimensional representation was computed using t-SNE, and the points representing the latent codes were colored according to the mean cell surface (a), volume (b), and sphericity (c). (d) shows a visual comparison of shapes with mean volume determined by the latent code. The latent codes were sampled along linear trajectories, where position (1) represents the lowest mean volume ($586.5$~{[\textmu$\text{m}^3$]}), position (3) the highest mean volume ($5719.9$~{[\textmu$\text{m}^3$]}), and position (5) the average volume ($3324.0$~{[\textmu$\text{m}^3$]})}
}
\label{figrev:latent_explore}
\end{figure*}


\begin{figure*}[!t]
\centering
\includegraphics[width=0.8\textwidth]{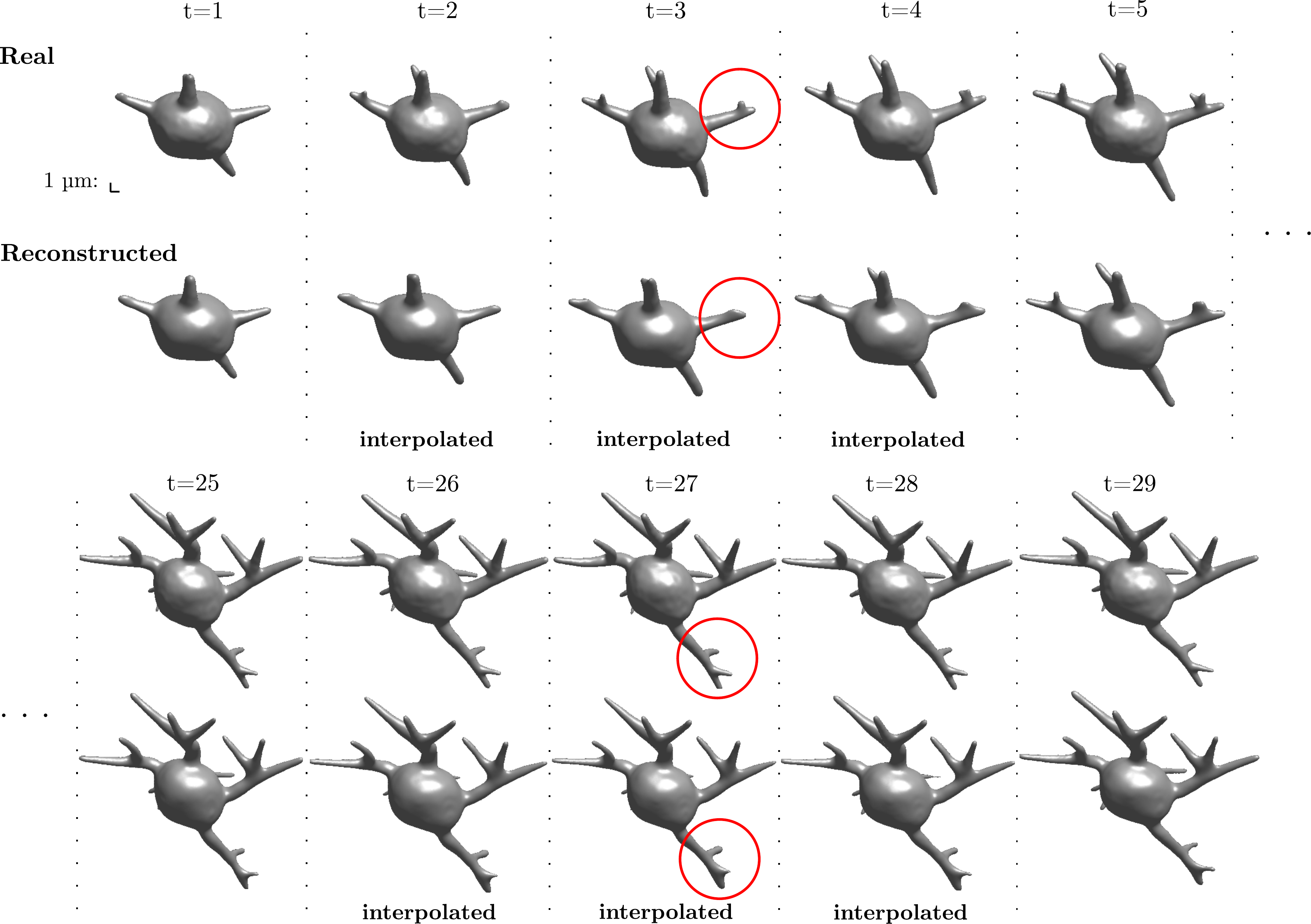}

\vspace{1mm}

\footnotesize{(a) Visual comparison of real and reconstructed shapes}

\vspace{2mm}

\includegraphics[width=0.45\textwidth]{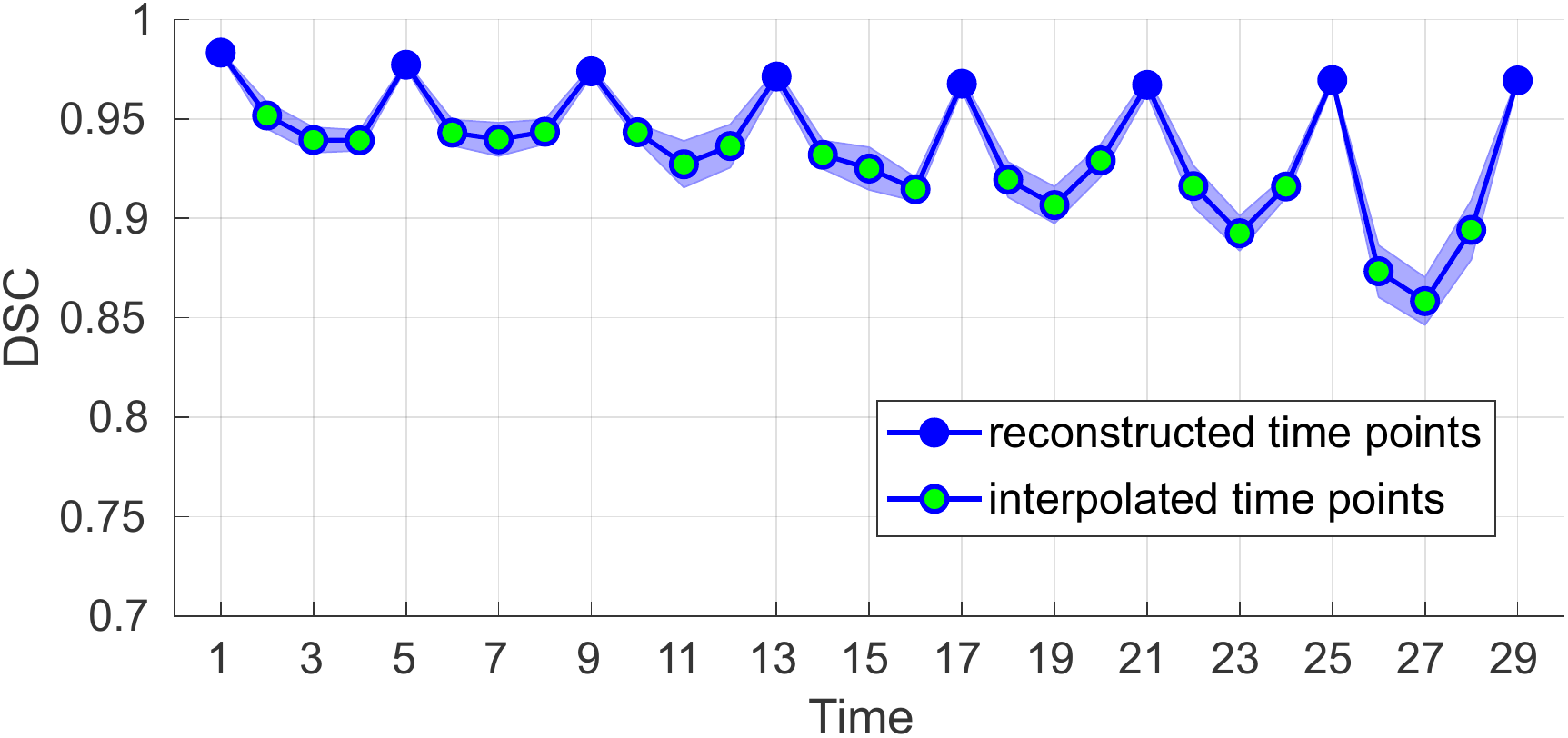}

\footnotesize{(b) Similarity of real and reconstructed shapes}

\vspace{-2mm}

\caption{\textbf{Temporal interpolation.} \textmd{We trained the proposed equivariant model on 33 time-evolving shapes in the $\mathcal{D}_{Filo}^{SDF}$ data set, with each sequence limited to every fourth time point. Specifically, we trained on time points $1, 5, 9,..., 29$, making a total of 8 time points per time series. (a) shows a visual comparison of real and reconstructed shapes on a single sequence, with the interpolated time points that the model has not seen explicitly marked as ``interpolated''. The differences are most apparent on the tips of the protrusions and are marked by a red circle. (b) shows DSC of real and reconstructed shapes computed over all sequences at each time point with the interpolated time points distinctly marked.}
}
\label{figrev:interpolation_sdf}
\end{figure*}

\subsection{Latent space compactness}
\label{sec:ablation}
In this experiment, we evaluated the impact of the SO(3)-equivariant extension on the compactness of the optimized latent space. In other words, we evaluated the ability of the equivariant model to learn a latent space that is rotation independent. The equivariant extension is depicted in blue in Fig.~\ref{fig:diagram_model}. We trained the \textit{non}-equivariant model~\citep{2022wiesner} and the proposed equivariant model on a set of randomly rotated time-evolving cell shapes. For this experiment, we selected the $\mathcal{D}_{Plat}^{SDF}$ data set. To prepare the training data, we pre-computed 4 random rotations of 25 time-evolving cell shapes from $\mathcal{D}_{Plat}^{SDF}$ to obtain a training set with 100 time-evolving shapes. Specifically, for each rotation, we uniformly sampled angles $\alpha, \beta$, and $\gamma$ from $(-\pi, \pi]$, and rotated the cell shape around the axis $x$, $y$, and $z$, respectively. Please note that the same angles were used at all time points for a given time-evolving shape. The dimensionality of latent codes was set to 64 for both models, as in the other experiments. Furthermore, the number of point samples, sampled randomly from the training data set at each epoch, was decreased to 250,000, and the batch size was increased to 20 to allow the model to observe more shapes at the same time and better optimize their rotations.

After training both models, we evaluated the resulting latent spaces. Each model learned a mapping of 100 time-evolving shapes to 100 locations in the respective latent space.
For visual comparison, we computed a low-dimensional representation of the latent spaces using t-SNE and plotted the resulting points in a scatter plot. Fig.~\ref{figrev:latent_plat:noeq} shows the latent space of the \textit{non}-equivariant model, and Fig.~\ref{figrev:latent_plat:eq} shows the latent space of the equivariant model. Rotated variants of each cell are represented by a unique marker. The plots show that the rotated variants of each cell are close together in the latent space of the equivariant model. 
Furthermore, we computed principal component analysis (PCA) of the latent codes learned by the models and then plotted the variability expressed by the individual principal components in Fig.~\ref{figrev:latent_plat2:pca}. The equivariant model exhibits $99.9\%$ variability in the first 20 principal components compared to $61.2\%$ of the \textit{non}-equivariant model. In other words, PCA shows that it may be possible to further reduce the dimensionality of the latent codes for the equivariant model. To evaluate this hypothesis, we trained both models on the same data set of rotated cells with latent code dimensionalities ranging from 4 to 128. Fig.~\ref{figrev:latent_plat2:dsc} and Table~\ref{tab:latent_plat2} show that the equivariant model retained the reconstruction similarity with latent dimensionality set to higher or equal to 32 (DSC of $0.954\pm0.009$ with dimensionality 32), whereas the \textit{non}-equivariant model required latent dimensionality of 64 or higher to achieve comparable results (DSC of $0.959\pm0.015$ with dimensionality 64, and $0.934\pm0.027$ with dimensionality 32). This experiment provides empirical evidence that the equivariant model is able to reconstruct shapes with higher similarity when the latent code dimensionality is reduced and supports the outcome of the PCA analysis.

\subsection{Latent space exploration}
In this experiment, we took a closer look at the latent space of the proposed equivariant model in order to assess its properties and determine how different positions in the latent space, represented by the latent codes, affect the resulting cell shape. We evaluate the latent space of the proposed equivariant model trained on 33 time-evolving shapes in $\mathcal{D}_{PLat}^{SDF}$. The training procedure yielded 33 latent codes with dimensionality of 64, representing a low dimensional embedding of the time-evolving shapes in $\mathcal{D}_{PLat}^{SDF}$. The standard deviation of the resulting latent codes was $0.0008$. To visualize the latent space and the positions of the latent codes, we computed a low dimensional representation using t-SNE. Fig.~\ref{figrev:latent_explore:surf}, Fig.~\ref{figrev:latent_explore:vol}, and Fig.~\ref{figrev:latent_explore:sph} show scatter plots containing markers representing the positions of individual cells in the latent space with points colored according to the shape descriptors that were computed on the cells, specifically, surface, volume, and sphericity, respectively. As each point in the latent space represents a time-evolving cell at multiple time points, we colored the points according to the mean descriptor values computed over all time points.  The plots show that different parts of the learned latent space correspond to cells with different surface, volume, and sphericity characteristics. In Fig.~\ref{figrev:latent_explore:sdf}, we show how different positions in the latent space affect the mean volume of a cell. We sampled latent codes along a linear trajectory, forming a triangle. Here, (1), (3), and (5) correspond to shapes that the model has seen during optimization. (1) represents a cell with the lowest mean volume ($586.5$~{[\textmu$\text{m}^3$]}), (3) a cell with the highest mean volume ($5719.9$~{[\textmu$\text{m}^3$]}), and (5) a cell with an average volume ($3324.0$~{[\textmu$\text{m}^3$]}). (2), (4), and (6) are sampled along a linear trajectory between the other ones. The visual comparison shows that the position in the latent space indeed determines the cell volume and that sampling latent codes along a linear trajectory results in a proper change in the cell volume.

\subsection{Temporal interpolation}
As the proposed shape representation is continuous, the model can be used to produce time-evolving shapes at arbitrary spatial and temporal resolution without the need for additional training. In this experiment, we evaluated the ability of the model to interpolate in time.
We trained the proposed equivariant model on every fourth time point present in the $\mathcal{D}_{Filo}^{SDF}$ data set (i.e., $1, 5, 9, ..., 29$), each time-evolving cell was thus represented by 8 time points. In other words, the model has only ``seen'' 8 time points for each of the 33 cells present in the data set. We subsequently used the model to fill in missing time points and compared the result to the real shapes. As time coordinates used for training and inference are linearly spaced in the interval from -1.0 to 1.0, representing the first and last time point, respectively, we shortened all sequences from 30 to 29 frames by discarding the last frame to ensure that the time coordinates are aligned. Fig.~\ref{figrev:interpolation_sdf}a shows a visual comparison of real and interpolated cell shapes on a single sequence. As the visual differences are difficult to discern with the naked eye, we computed the similarity of the shapes using DSC, which is shown in Fig.~\ref{figrev:interpolation_sdf}b. The plot shows the mean and standard deviation of DSC computed over all sequences at each time point. The DSC shows that the interpolation similarity decreases toward the end of the sequence, where the cell protrusions are fully grown, and the shape is most complex. The interpolation yielded the lowest DSC of $0.858\pm0.012$ at time point 27. The visual comparison shows that the model is able to interpolate between time points, and DSC corroborates that the produced shapes retain reasonable similarity to the real cells even when interpolating multiple consecutive time points.

\subsection{Spatial interpolation and sampling grid}
In this experiment, we investigated how the size of the sampling grid affects the shape reconstruction. We trained the proposed equivariant model with shape SDFs represented by three different grids: $192\!\times\!192\!\times\!192$, $256\!\times\!256\!\times\!256$, and $384\!\times\!384\!\times\!384$. As the original microscopy images of A549 filopodial cells are isotropic and exhibit high spatial resolution of $300\!\times\!300\!\times\!300$ voxels, we chose shapes of this cell line for this experiment. We already had the $\mathcal{D}_{Filo}^{SDF}$ data set consisting of $256\!\!\times\!\!256\!\!\times\!\!256$ point samples per time point. To prepare suitable training data sets for the other two models, we computed SDFs and created two additional variants of the data set, $\mathcal{D}_{Filo}^{SDF\_192}$ and $\mathcal{D}_{Filo}^{SDF\_384}$, with sampling grids of $192\!\times\!192\!\times\!192$, and $384\!\times\!384\!\times\!384$ points, respectively. We will refer to resulting models as $\texttt{model\_192}$, $\texttt{model\_256}$, and $\texttt{model\_384}$, with respect to the grid used for training.

\begin{figure*}[!p]
\centering
    \captionsetup[subfigure]{justification=centering} 
    
    \begin{subfigure}[b]{0.95\textwidth}
        \centering
        \includegraphics[width=\textwidth]{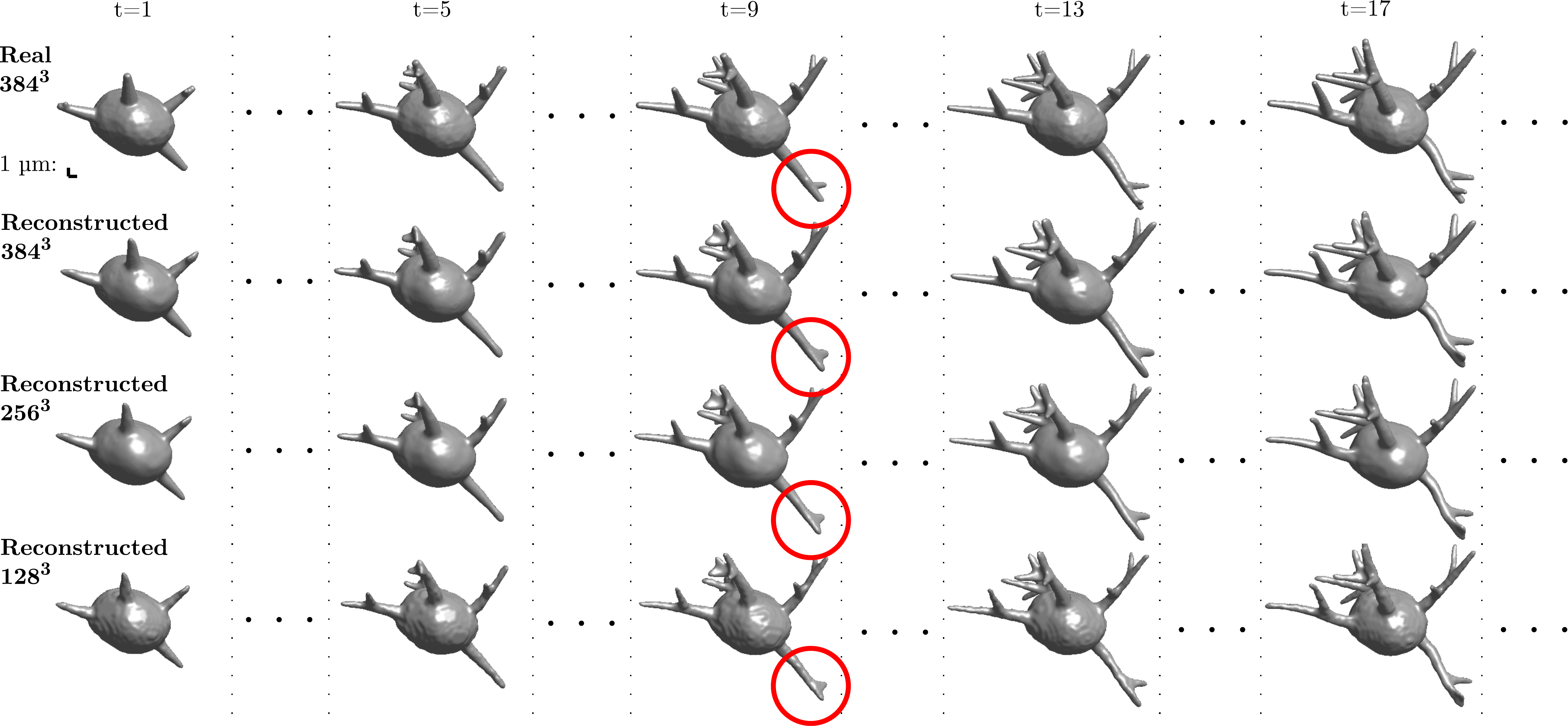}
        \caption{Model trained on grid of $384\!\times\!384\!\times\!384$ samples}
        \label{figrev:interpolation_spatial:384}
    \end{subfigure}

    \begin{subfigure}[b]{0.95\textwidth}
        \centering
        \includegraphics[width=\textwidth]{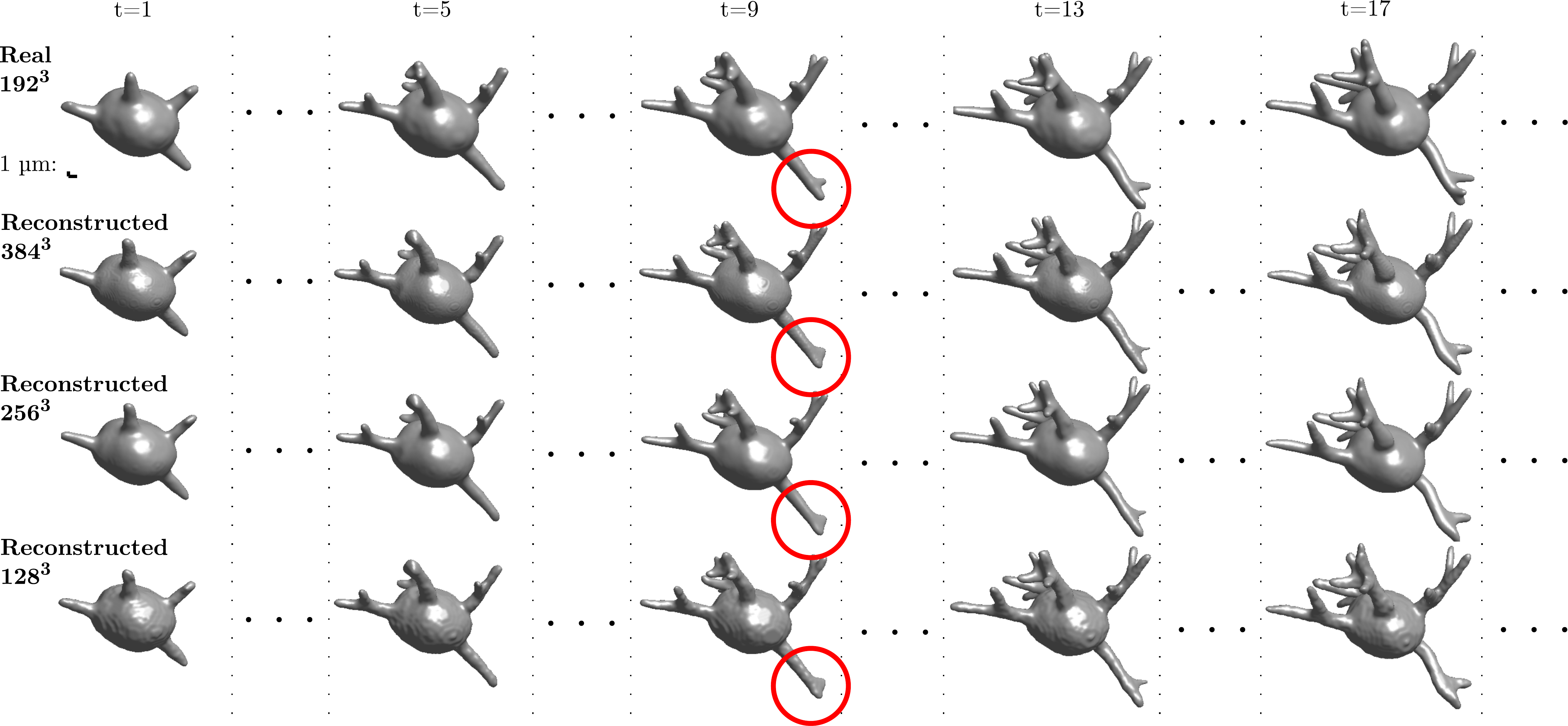}
        \caption{Model trained on grid of $192\!\times\!192\!\times\!192$ samples}
        \label{figrev:interpolation_spatial:192}
    \end{subfigure}

    \begin{subfigure}[b]{0.4\textwidth}
        \centering
        \includegraphics[width=\textwidth]{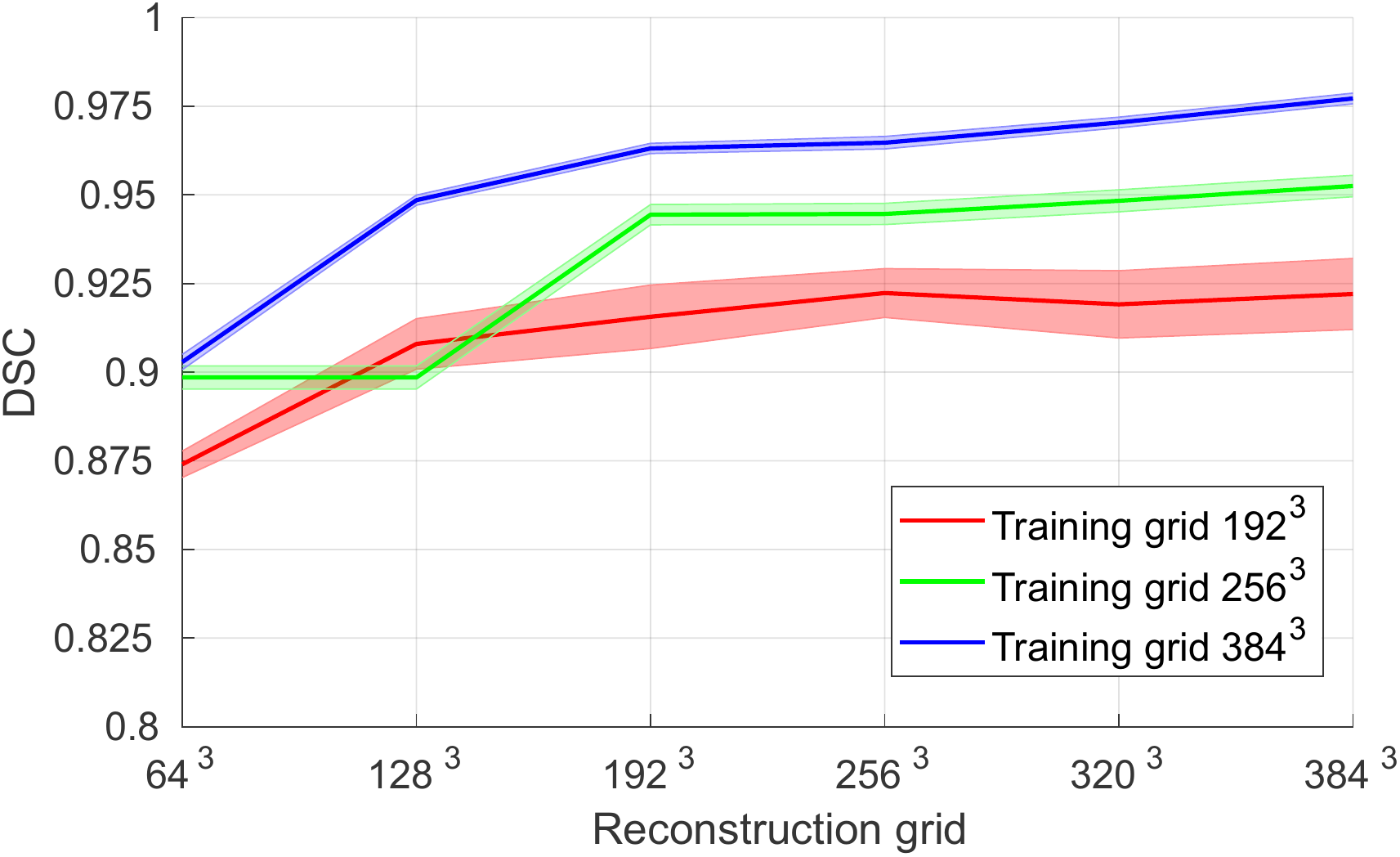}
        \caption{Dice similarity coefficient with respect to the sampling grid}
        \label{figrev:interpolation_spatial:dice}
    \end{subfigure}
    
\vspace{-2mm}

\caption{\textbf{Reconstructed shapes with respect to the training and reconstruction grids.} \textmd{The proposed equivariant model was trained on $\mathcal{D}_{Filo}^{SDF}$ with a grid of $192^3$, $256^3$, and $384^3$. Subsequently, the model was used to reconstruct cell shapes with different grids. Specifically, $384^3$, $320^3$, $256^3$, $192^3$, $128^3$, and $64^3$. Visualization of the reconstruction results showing selected time points of a single time evolving cell is shown for the model trained with a grid $384^3$ (a) and grid $192^3$ (b). The differences are most prominent on the tips of the protrusions, which are marked by a red circle. The Dice similarity coefficient (DSC) of real and reconstructed shapes with respect to the training and reconstruction grid is shown in (c).}
}
\label{figrev:interpolation_spatial}
\end{figure*}

After training the models, we reconstructed the cell shapes using different grids, ranging from $64^3$ to $384^3$. As the shape representation with the proposed model is continuous, the model can be used to reconstruct shapes using an arbitrary grid, isotropic or anisotropic, in other words, to interpolate in space. Fig.~\ref{figrev:interpolation_spatial:384} and Fig.~\ref{figrev:interpolation_spatial:192} show a visual comparison of shapes reconstructed using $\texttt{model\_384}$ and $\texttt{model\_192}$, respectively. As the visual differences may be difficult to spot with the naked eye, we also computed DSC for reconstructed cells, including all three models and grid sizes, in Fig.~\ref{figrev:interpolation_spatial:dice}. DSC was computed over all sequences at time point 30, where the cell protrusions are fully grown. All Reconstructed cell shapes at the given time point were compared against real shapes with the grid of $384^3$. Voxel volumes of shapes reconstructed with smaller grids were upsampled using the nearest neighbor algorithm to match the $384^3$ grid for DSC computation.

The reconstruction with grid of $384^3$ yields the best DSC on all models. The results show that it is possible to further increase the reconstruction similarity by training the model on SDFs with denser grids, as demonstrated by $\texttt{model\_384}$ trained on $\mathcal{D}_{Filo}^{SDF\_384}$. Using denser training grids does not affect memory consumption, which is determined only by the batch size and the number of points sampled in each epoch. However, the grid density affects the number of epochs that the models need to converge. Specifically, $\texttt{model\_192}$ required 1000 epochs, $\texttt{model\_256}$ 2000 epochs, and $\texttt{model\_384}$ needed 4000 epochs.

\begin{figure*}[!t]
\centering
\captionsetup[subfigure]{justification=centering} 
    
    \begin{subfigure}[b]{0.95\textwidth}
        \centering
        \includegraphics[width=\textwidth]{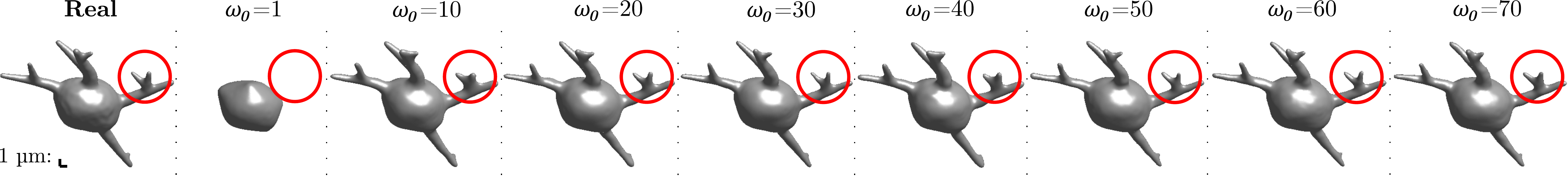}
        \caption{Visual comparison of shapes reconstructed with different values of the $\omega_0$ parameter}
        \label{figrev:omegasdf:sdf}
    \end{subfigure}

    \begin{subfigure}[b]{0.45\textwidth}
        \centering
        \includegraphics[width=\textwidth]{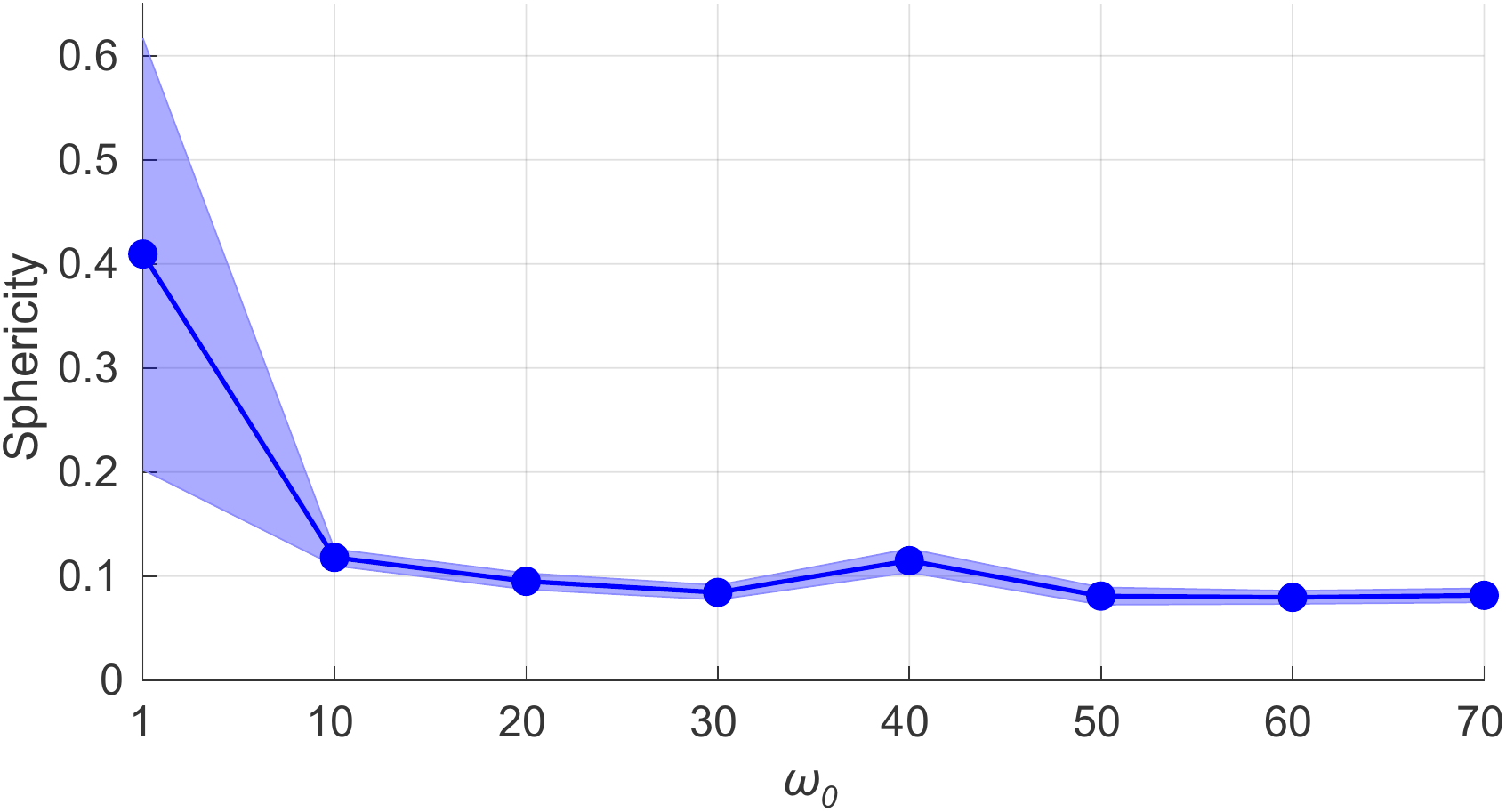}
        \caption{Sphericity of shapes reconstructed with different values of the $\omega_0$ parameter}
        \label{figrev:omegasdf:sph}
    \end{subfigure}

\vspace{-2mm}

\caption{\textbf{Comparison of reconstructed shapes with respect to the $\bm\omega_0$ parameter.} \textmd{We trained the proposed equivariant model on the $\mathcal{D}_{Filo}^{SDF}$ data set with $\omega_0$ parameter ranging from $1$ to $70$. (a) shows a visual comparison of real and reconstructed shapes on a single cell at time point 30 where the protrusions are fully grown. The differences between real and reconstructed shapes are most apparent on the tips of the protrusions and are marked by a red circle. (b) shows the mean and standard deviation of sphericity computed at time point 30 over all reconstructed shapes with respect to the $\omega_0$ parameter.}
}
\label{figrev:omegasdf}
\end{figure*}

\begin{figure*}[!t]
\centering
\captionsetup[subfigure]{justification=centering} 
    
    \begin{subfigure}[b]{\textwidth}
        \centering
        \includegraphics[width=\textwidth]{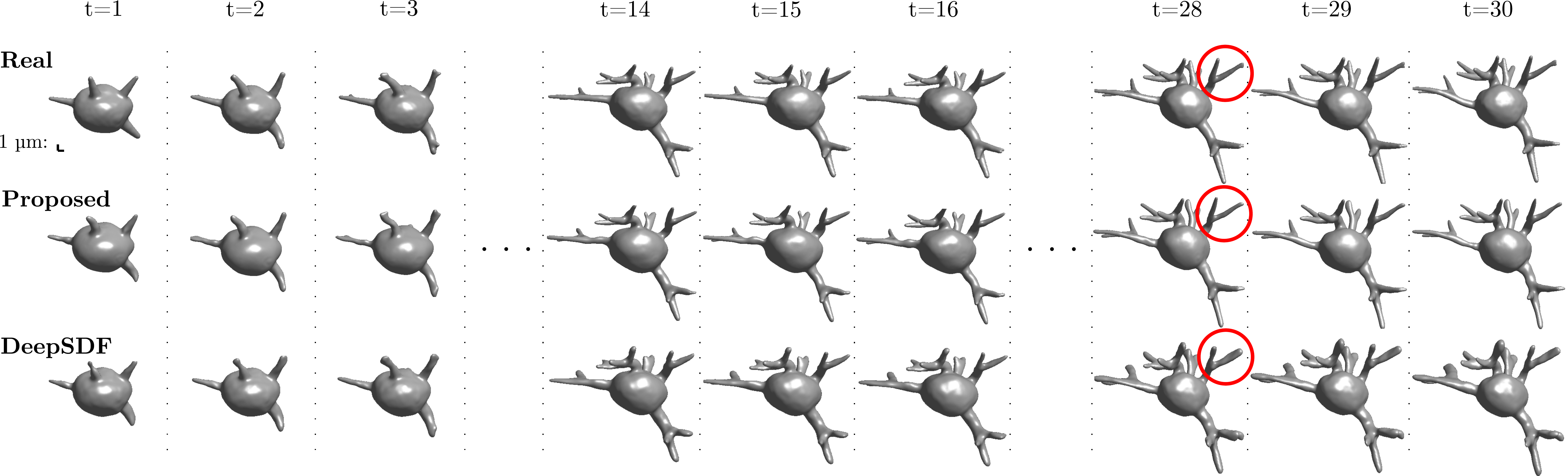}
        \caption{Visual comparison of real and reconstructed shapes}
        \label{figrev:sine_relu:sdf}
    \end{subfigure}

    \begin{subfigure}[b]{0.4\textwidth}
        \centering
        \includegraphics[width=\textwidth]{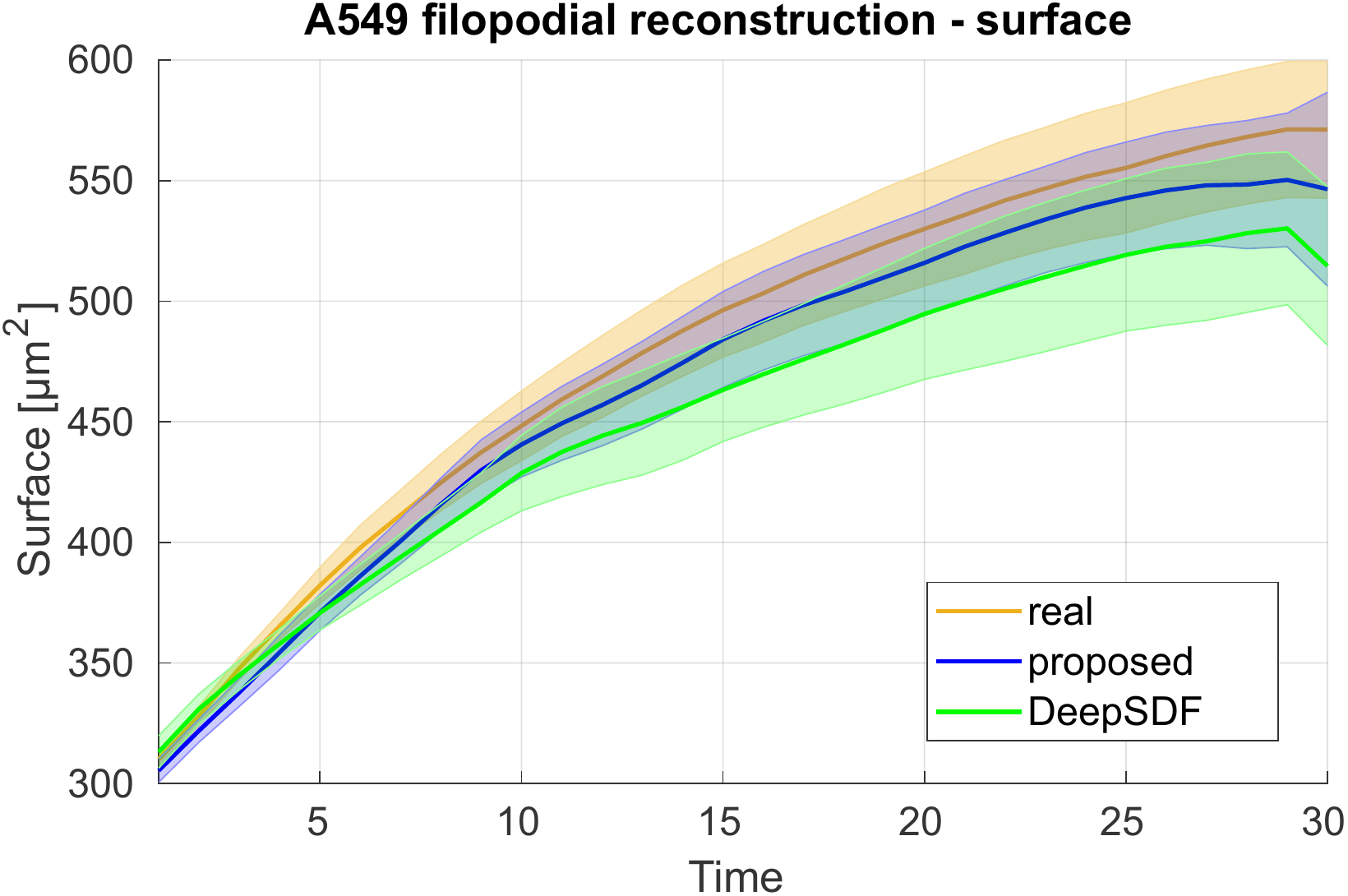}
        \caption{Shape surface over time}
        \label{figrev:sine_relu:surf}
    \end{subfigure}
    \hspace{2.7cm}
    \begin{subfigure}[b]{0.4\textwidth}
        \centering
        \includegraphics[width=\textwidth]{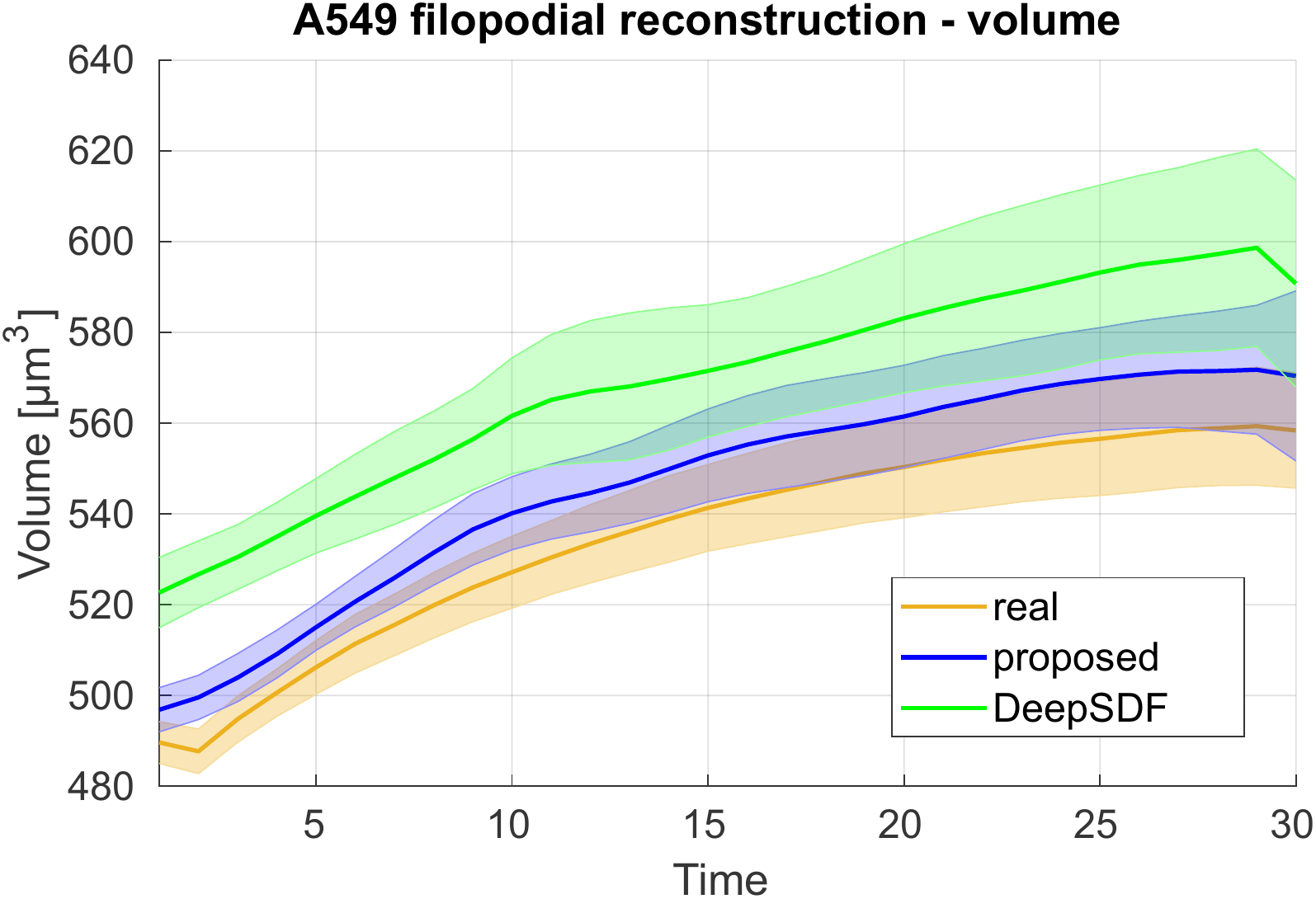}
        \caption{Shape volume over time}
        \label{figrev:sine_relu:vol}
    \end{subfigure}

    \begin{subfigure}[b]{0.4\textwidth}
        \centering
        \includegraphics[width=\textwidth]{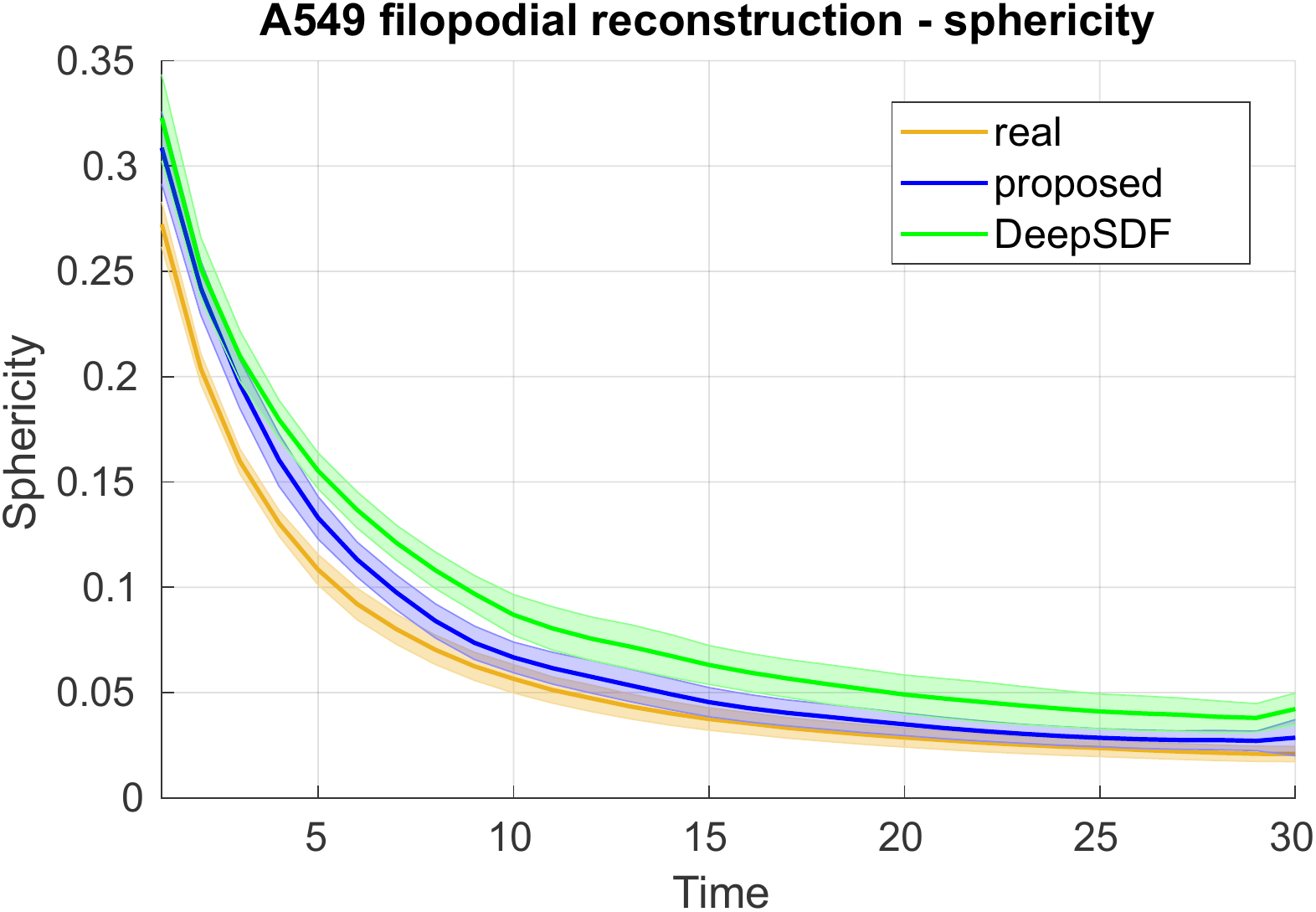}
        \caption{Shape sphericity over time}
        \label{figrev:sine_relu:sph}
    \end{subfigure}
    \hspace{1cm}
    \begin{subfigure}[b]{0.495\textwidth}
        \centering
        \includegraphics[width=\textwidth]{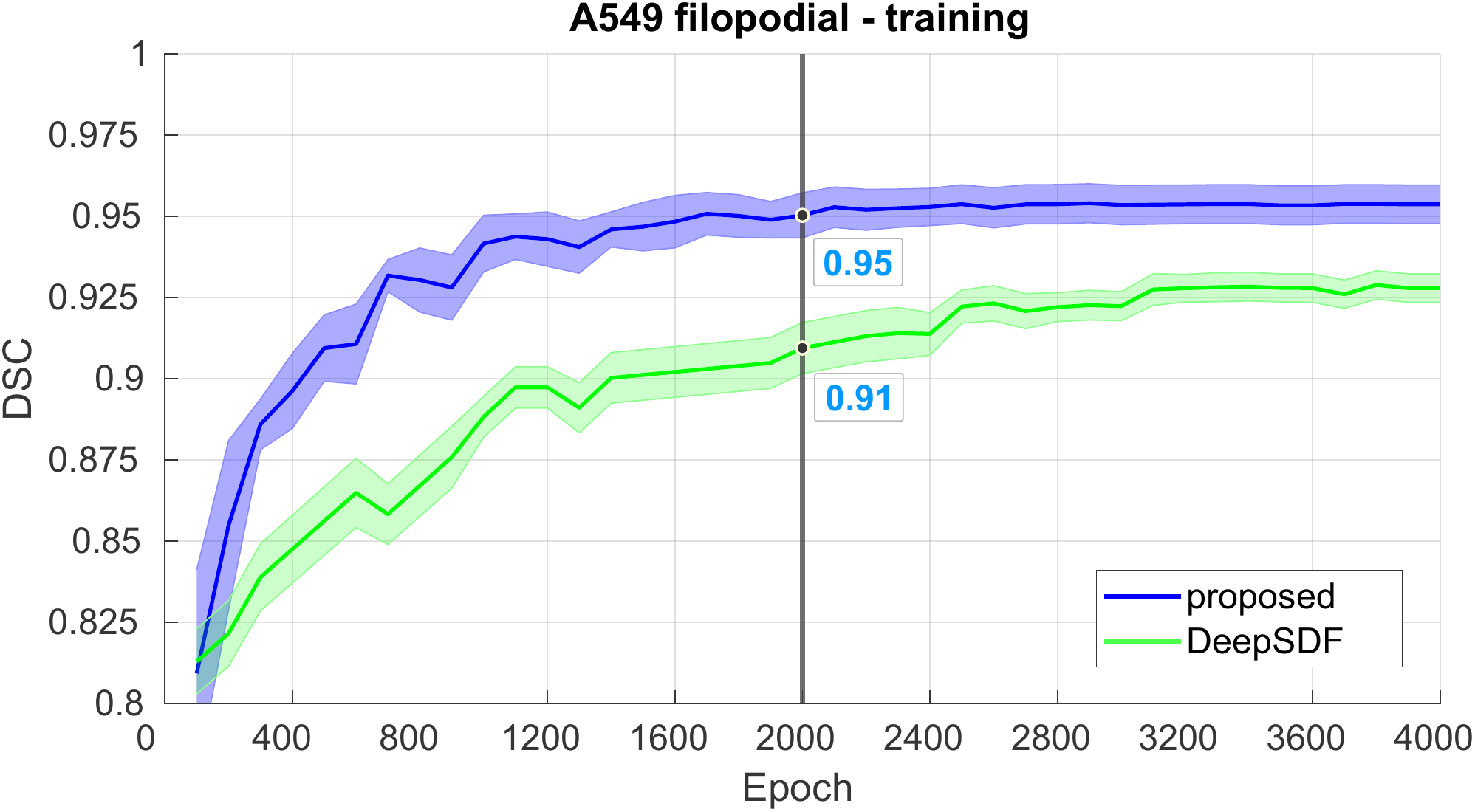}
        \caption{Dice similarity coefficient (DSC) of real and reconstructed shapes}
        \label{figrev:sine_relu:conv}
    \end{subfigure}

    \vspace{-2mm}
    
\caption{\textbf{Comparison of the proposed equivariant model with DeepSDF.} \textmd{Both models were trained on the $\mathcal{D}_{Filo}^{SDF}$ data set for 4000 epochs. (a) shows a visual comparison of real and reconstructed shapes on a single sequence. The differences between the shapes are most apparent on the protrusions (red circle), where the shapes reconstructed using the proposed model are sharper and more faithful to the real ones. The plots (b), (c), (d) show mean and standard deviation of surface, volume, and sphericity, respectively, at each time point over all sequences. Where ``real'' represents real cells from the training data set, ``proposed'' represents the shapes reconstructed using the equivariant model, and ``DeepSDF'' represents the shapes reconstructed using DeepSDF. Additionally, we compare the convergence of the models using Dice similarity coefficient (DSC) (e). DSC of real and reconstructed shapes was computed every 100th epoch over all sequences at the last time point (30), where the cell protrusions are fully grown.}
}
\label{figrev:sine_relu}
\end{figure*}


\subsection{Spectral bias}
In this experiment, we investigate how the $\omega_0$ parameter affects the spectral bias of the model. The weights of layers using periodic activations are initialized using a parameter $\omega_0$ representing the angular frequency of the periodic functions. This parameter was shown to affect the spectral bias of the neural network in~\citep{sitzmann2020implicit}, and its recommended value was 30, which is the value that we used in our other experiments. To investigate how the $\omega_0$ parameter affects the shape reconstruction of the proposed equivariant model, we trained multiple models on the $\mathcal{D}_{Filo}^{SDF}$ data set with values of $\omega_0$ ranging from 1 to 70. Fig.~\ref{figrev:omegasdf:sdf} shows a visual comparison of a real cell shape from the training data set and the shapes reconstructed with models trained with different $\omega_0$ values. For visualization, we selected a single shape at time point 30 where the cell protrusions are fully grown, and the cell exhibits the lowest sphericity. The differences are most apparent on the tips of the protrusions, where the models trained with $\omega_0$ lower than 30 produce shapes with a rounder and less defined structure. On the other hand, increasing $\omega_0$ beyond 30 did not seem to yield any visually observable improvements over the default recommended value. 
Fig.~\ref{figrev:omegasdf:sph} shows a plot of sphericity with respect to different $\omega_0$ values computed over all reconstructed cells at time point 30. The results indicate that the $\omega_0$ parameter indeed affects the spectral bias of the model and that increasing this parameter beyond the value recommended by \citep{sitzmann2020implicit} does not yield any measurable improvements in sharpness of the reconstructed shapes.

\begin{figure}[!t]
\centering
\includegraphics[width=\columnwidth]{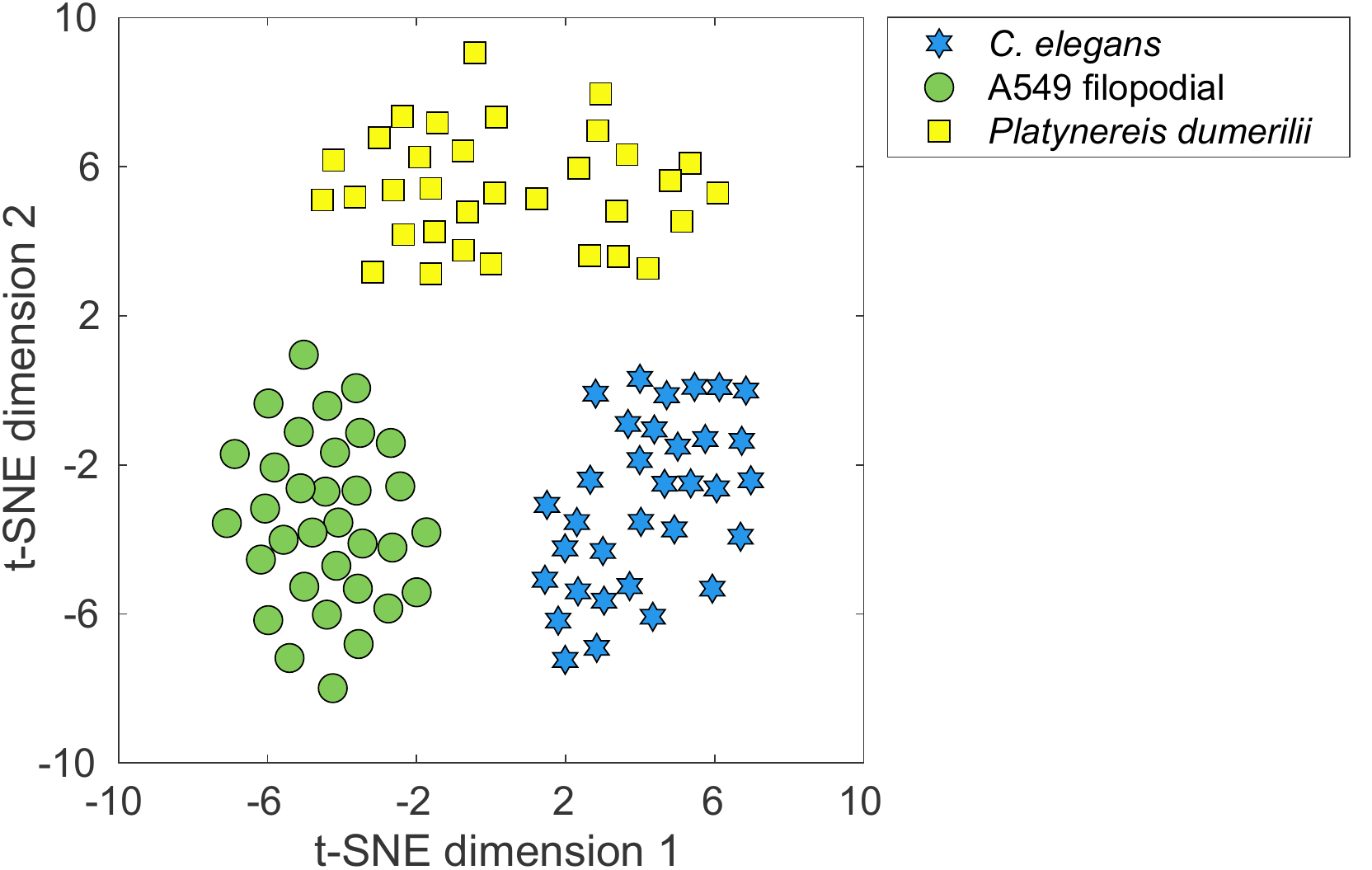}
\caption{\textbf{Cell classification based on a latent code.} \textmd{The figure shows a low dimensional representation of the learned latent space computed using t-SNE. The model was trained on three cell lines, specifically, \textit{Platynereis dumerilii} cells, \textit{C. elegans} cells, and A549 filopodial cells. Each cell line is represented by 33 time-evolving cell shapes, making a total of 99 cells over all three cell lines. Time-evolving cell shapes of each line form separate, distinct clusters in the latent space.}}
\label{figrev:latclas}
\end{figure}

\begin{figure*}[!t]
\centering
\includegraphics[width=\textwidth]{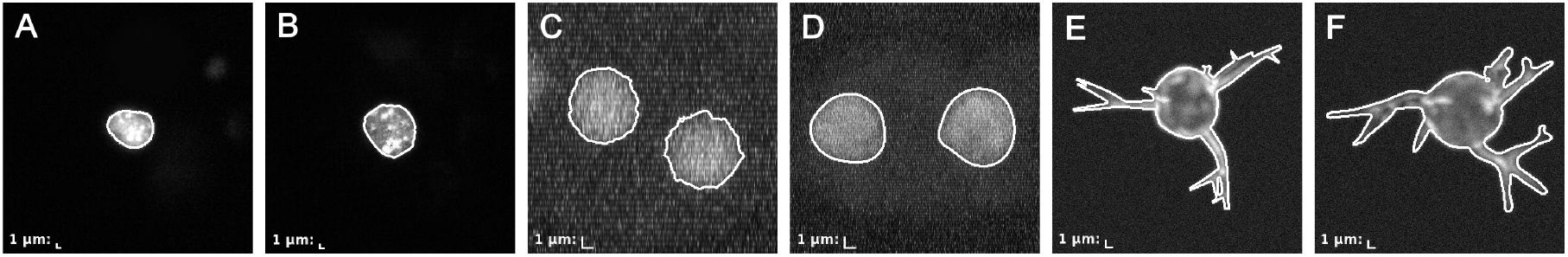}
\caption{\textbf{Comparison of real (A, C, E) and synthetic (B, D, F) images of \textit{Platynereis dumerilii} (A, B), \textit{C.~elegans} (C, D), and A549 filopodial cells (E, F).} \textmd{The images show one frame from the respective 2D time-lapse data sets. The segmentation masks are represented as white contours. The masks were obtained using the proposed method and the texture was produced using an image-to-image model.}}\label{fig:benchmark}
\end{figure*}

\subsection{Comparison with DeepSDF}
In this experiment, we compared the proposed equivariant model with DeepSDF~\citep{park2019deepsdf}. DeepSDF differs from the \textit{non}-equivariant and the equivariant models presented in this study in its architecture. Specifically, DeepSDF uses ReLU activation functions, and the MLP is composed of 8 layers. In comparison, our models have MLP with 9 layers and use \textit{sine} activation functions, where the equivariant model is further extended to be rotation equivariant. Periodic activation functions (\textit{sine}) proposed by~\citep{sitzmann2020implicit} are expected to yield better results in comparison to ReLU activations. Specifically, they should allow the model to converge faster and to better fit high-frequency signals, such as shapes with sharp edges. In order to compare the methods, we used a DeepSDF extended to model shapes in 3D+time~\citep{2022wiesner}.

We experimented on the $\mathcal{D}_{Filo}^{SDF}$ data set, which consists of 33 complex cell shapes with sharp growing and branching protrusions. Furthermore, we extended the number of training epochs to 4000 to give both models enough time to converge and set the latent code dimensionality of DeepSDF to 256~\citep{park2019deepsdf}. We evaluated the results of both models at epoch 4000. A visual comparison of the real and resulting reconstructed cell shapes is shown in Fig.~\ref{figrev:sine_relu:sdf}. The differences are most apparent on the cell protrusions toward the end of the sequence when the cell is fully grown. A visual inspection of the shapes shows that the reconstruction using DeepSDF yields thicker protrusions with rounder tips compared to the proposed model with periodic activations that was able to match the real shapes more closely. We show quantitative evaluation using shape descriptors, specifically, surface in Fig.~\ref{figrev:sine_relu:surf}, volume in Fig.~\ref{figrev:sine_relu:vol}, and sphericity in Fig.~\ref{figrev:sine_relu:sph}, and the training convergence in Fig.~\ref{figrev:sine_relu:conv}. The quantitative results support the visual inspection, with DeepSDF producing rounder shapes that exhibit higher volume due to the increased thickness of the protrusions. At time point 30 where the cell is fully grown (and exhibits the lowest sphericity), DeepSDF and the proposed model yielded sphericity $0.042\pm0.008$ and $0.029\pm0.009$, respectively, compared to real shapes with $0.021\pm0.004$. To show the training convergence, we computed DSC of real and reconstructed shapes every 100th epoch over all sequences at time point 30. The proposed model was able to converge towards a DSC of $0.950\pm0.007$ in 2000 epochs, whereas DeepSDF yielded DSC of $0.909\pm0.008$ at epoch 2000 and needed another 2000 epochs to converge toward DSC of $0.928\pm0.004$. The results show that the reconstruction similarity and the convergence speed of the proposed equivariant model are measurably improved in comparison to DeepSDF.

\subsection{Cell classification based on a latent code}
In this experiment, we investigated the application of the proposed equivariant model in a downstream classification task. We optimized a single model on all three data sets, specifically, $\mathcal{D}_{Plat}^{SDF}$, $\mathcal{D}_{Cele}^{SDF}$, and $\mathcal{D}_{Filo}^{SDF}$. Each data set consists of 33 time-evolving cell shapes of a given cell line, making a total of 99 cells. We visualized the latent space of the model using t-SNE in Fig.~\ref{figrev:latclas}. The shape time series of each cell line form a distinct cluster in the latent space. The model distinguishes between different shape features of each cell line and puts similar shapes close together in the latent space. On data sets where the cell class labels are not known, the cells can be labeled according to their position in the latent space.

\subsection{Conditional synthesis of textured cell images}
To demonstrate the application of the method, we used synthetic time-evolving shapes (see Sec.~\ref{sec:reconstruction}) to generate synthetic data sets with pairs of a cell shape mask and a corresponding microscopy image. A comparison of real and synthetic microscopy images is shown in Fig.~\ref{fig:benchmark}. Such data sets are used as training data for segmentation networks or for testing and evaluation of image analysis methods. In the latter case, we refer to them as \textit{benchmarking} data sets.

The segmentation masks were obtained using maximum intensity projections of the voxel volumes of SDFs produced using the proposed method. The textured cell images were generated using a conditional GAN, more specifically pix2pixHD~\citep{2018wang}, which was trained on microscopy images and corresponding cell masks of \textit{Platynereis dumerilii} cells, \textit{C. elegans} cells, and A549 human carcinoma cells. The resulting data sets are produced in 2D+time and contain 33 time-lapse sequences with 30 time points for each cell type.

\section{Discussion}
In this work, we have proposed a generative model for living cell shapes in 3D+time. We represent evolving cell shapes using the zero-level set of their signed distance function, which is implicitly represented in a fully-connected neural network. This implicit neural representation is fully continuous and thus allows the synthesis of highly detailed shapes in virtually unlimited spatial and temporal resolution. By disentangling shape from rotation, we obtain a compact latent code that allows reconstruction and synthesis of cells with diverse shape and growth characteristics. In a series of experiments, we have shown that this model can be used to accurately reconstruct, synthesize, and interpolate complex and changing cell shapes. 

In the proposed model, shape and rotation are disentangled, so that the latent space describes only shape. This has several advantages. First, we showed that this results in learning a more compact latent space, as fewer latent space dimensions might be necessary to describe a data set. Second, by explicitly specifying the rotation, the model is capable of generating cell shapes at any desired angle in 3D space. Third, by allowing the model to focus on the optimization of a latent description of shape, it might be able to learn a latent space in which similar cell shapes are properly clustered. This might allow unsupervised learning based on cell shapes, clustering, and identification of categories of cells. As the latent code describes shape changes over time, it is likely to also capture differences in morphology between cells, and provide insights into cell development which -- in turn -- can be used for deriving accurate quantitative models, e.g., for embryogenesis. Such flexibility and level of detail would not have been possible with existing voxel-based methods~\citep{Svoboda:MitoGen2017, 2018fu, 2019baniukiewicz}, nor in our previous work employing implicit neural representations~\citep{2022wiesner} 

In our experiments, we have included three diverse data sets to demonstrate the versatility of our approach: \textit{Platynereis dumerilii} embryo cells, \textit{C. elegans} embryo cells, and A549 lung adenocarcinoma cancer cells. The \textit{Platynereis dumerilii} and \textit{C. elegans} embryo cells are commonly used in biological evolution studies as model organisms. The A549 lung adenocarcinoma cancer cells are used in cancer studies and are subject to active research because filopodia and their relationship to cell migration are of great importance to the testing and development of drug therapies and understanding of the formation of cancer metastases. Our experiments showed that the model is able to accurately represent diverse time-evolving cell shapes and the phenomena occurring during the cell cycle, such as growth and mitosis. In general, descriptive statistics of synthesized cells matched those of real cells. In particular, the surface and volume of real and synthetic cells followed the same distribution. However, we found that the sphericity of synthetic cells was generally higher than that of real cells, indicating that it might still be challenging to properly synthesize details on the surface. A possible mitigation strategy for this might be to allow adaptive shape sampling. One can use sparse sampling where the shape is very smooth, and dense sampling where the shape contains fine details, or simply deserves more attention and thus accuracy, like protrusions on a cell surface. Similary, we can adaptively sample in time, where high temporal resolution has been shown to improve segmentation and tracking results on time-lapse data~\citep{2014coca}. 

By random sampling in the optimized latent space, we were able to generate new time-evolving cell shapes with visually plausible features. This random sampling produced consistent results when synthesizing new \textit{Platynereis dumerilii} and \textit{C. elegans} cells. However, in the case of the A549 filopodial cells, we observed that the fully-grown protrusions of the new synthetic shapes had a tendency to disconnect from the main cell body in the second half of the time-lapse sequence. To mitigate this phenomenon, we randomly sampled new latent codes close to known latent codes of our training data. Additional regularization terms could help with optimizing a latent space that would be more suitable for sampling complex and heterogenous time-evolving shapes, such as the A549 filopodial cells. It's also possible that the complexity of this data set requires a more complex latent space distribution to sample new shapes. In future work, we will investigate how more structure and guarantees can be obtained for the latent space and explore Gaussian mixture models~\citep{reynolds2009gaussian}, which are able to generate more complex data distributions than a single isotropic Gaussian.

A limitation of the current study is that the A549 human lung carcinoma cells that we used to optimize our synthesis model were themselves synthesized~\citep{sorokin2018filogen} and thus might be slightly different from real living cells. We argue that this is only a minor limitation, as we use these data sets to demonstrate the efficacy of the model and show that our method can indeed synthesize cells that mimic the distribution of these reference cells. Moreover, we only model individual cells, while in reality, cells form populations. In future work, we wish to explore extensions to arrive at a 3D+time model capable of synthesizing living cell populations, describing not only the time-evolving shapes of cells but also cell trajectories and even cell interactions within the population. This would allow the synthesis of data sets that can be used for the development of instance segmentation or tracking models, which can distinguish between cells in a population.

Implicit neural representations are a versatile tool and, depending on the desired application, the inferred SDFs can be converted to mesh-based, voxel-based, or point cloud representations. To demonstrate one potential application of the proposed method, we used a GAN conditioned on the synthetic cell shapes to prepare data sets with pairs of textured cell images and reference annotation for all three cell types. This is a similar approach to those previously used for synthesis of new data~\citep{2017osokin,2017goldsborough,2019bohland,2019bailo,2019baniukiewicz,2021kozlovsky}, and the acquired data sets could be valuable for training, evaluation, and benchmarking of image analysis algorithms. In this study, we demonstrated the proposed method on evolving cell shapes from optical microscopy, but in principle, this approach can be used for learning shapes and spatio-temporal dynamics of diverse organisms at both micro and macro scales. For example, this method could be adapted to synthesize brain atrophy in patients with Alzheimer's disease or the progression of abdominal aortic aneurysms~\citep{alblas2023implicit}.

\section{Conclusion}
We have presented a method that allows accurate spatio-temporal representation and synthesis of highly-detailed time-evolving shapes and structures in microscopy imaging. The method uses a neural network to implicitly represent time-evolving shapes and the occurring visual phenomena and deformations. This representation is fully continuous and equivariant with respect to shape rotations and allows for the synthesis of shapes in virtually unlimited spatial and temporal resolution at any given rotation.

In conclusion, conditional rotation equivariant implicit neural representations are a suitable representation for generative modeling of living cells.

\section*{Acknowledgments}
The data set of \textit{Platynereis dumerilii} embryo cells is courtesy of Mette Handberg-Thorsager and Manan Lalit, who both have kindly shared it with us.

The shape descriptors were computed and plotted using an online tool for quantitative evaluation, Compyda, available at \url{https://cbia.fi.muni.cz/compyda}. We thank its authors T. Ne\v{c}asov\'{a} and D. M\'{u}\v{c}ka for kindly giving us early access to this tool and facilitating the evaluation of the proposed method.

\bibliographystyle{model2-names.bst}\biboptions{authoryear}
\bibliography{refs}



\section*{Supplementary material}

# Generative modeling of living cells with SO(3)-equivariant implicit neural representations

## Supplementary material

David Wiesner[*1], Julian Suk[2], Sven Dummer[2], Tereza Nečasová[1],
Vladimír Ulman[3], David Svoboda[1], and Jelmer M. Wolterink[2]

[1] Centre for Biomedical Image Analysis, Masaryk University, Brno, Czech Republic
[2] Department of Applied Mathematics & Technical Medical Centre, University of Twente, Enschede, The Netherlands
[3] IT4Innovations, VSB – Technical University of Ostrava, Ostrava, Czech Republic
[*] Corresponding author: `wiesner@fi.muni.cz`

## 1 Overview

This supplementary material provides additional proofs of the equivariance of the method proposed in the main paper.

## 2 Equivariance proofs

In the main text, we introduce two loss functions that can be minimized to retrieve the rotation matrix $\boldsymbol{R}_\varphi$ and the latent code $\boldsymbol{z}$ of a shape time series $\mathcal{M}_t$: one loss function for the infinite sample case and one for the finite sample case. These are respectively given by:

$$\mathcal{L}_{\mathcal{M}_t}(\boldsymbol{R}_\varphi, \boldsymbol{z}) := \mathbb{E}_{(\boldsymbol{x},t) \sim \rho_{\mathcal{M}_t}} \bigg( \mathcal{L}_{rec}(f_\theta(\boldsymbol{R}_\varphi^T \boldsymbol{x}, t, \boldsymbol{z}), \mathrm{SDF}_{\mathcal{M}_t}(\boldsymbol{x})) + \mathcal{L}_{code}(\boldsymbol{z}) \bigg). \tag{S.1}$$

and

$$\mathcal{L}_{\mathcal{M}_t}^M(\boldsymbol{R}_\varphi, \boldsymbol{z}) := \frac{1}{M} \sum_{i=1}^{M} \bigg( \mathcal{L}_{rec}(f_\theta(\boldsymbol{R}_\varphi^T \boldsymbol{x}_i, t_i, \boldsymbol{z}), s_i) + \mathcal{L}_{code}(\boldsymbol{z}) \bigg). \tag{S.2}$$

where $\{(\boldsymbol{x}_i, t_i, s_i)\}_{i=1}^M$ are samples $s_i = \mathrm{SDF}_{\mathcal{M}_{t_i}}(\boldsymbol{x}_i)$ and $\rho_{\mathcal{M}_t} : \Omega \times [0,1] \to \mathbb{R}$ is a time dependent spatial distribution which can differ per $\mathcal{M}_t$. Here $\Omega$ is the domain in which the shape lives.

In this Appendix, we prove that for both losses we have an equivariant decoder:

**Theorem** (Equivariance of our model in the infinite sample limit). *Assume we rotate a shape time series $\mathcal{M}_t$ using a rotation matrix $\tilde{\boldsymbol{R}}$ and obtain shape time series $\mathcal{M}_t^{\tilde{\boldsymbol{R}}}$. Furthermore, assume that the reconstruction for a general shape time series is obtained by minimizing Equation S.1 over $(\boldsymbol{R}_\varphi, \boldsymbol{z})$ and assume without loss of generality that $\Omega$ is a sphere centered around the origin. Finally, assume that a solution to the optimization problem for $\mathcal{M}_t$ is $(\boldsymbol{R}, \boldsymbol{z})$ and that $\rho_{\mathcal{M}_t^{\tilde{\boldsymbol{R}}}}(\boldsymbol{x}, t) = \rho_{\mathcal{M}_t}(\tilde{\boldsymbol{R}}^T \boldsymbol{x}, t)$. Then, a solution of the optimization problem for $\mathcal{M}_t^{\tilde{\boldsymbol{R}}}$ is $(\tilde{\boldsymbol{R}}\boldsymbol{R}, \boldsymbol{z})$.*

*Proof.* First, note that the function we need to minimize is:

$$\mathcal{L}_{\mathcal{M}_t^{\tilde{\boldsymbol{R}}}}(\boldsymbol{R}_\varphi, \boldsymbol{z}) = \mathbb{E}_{(\boldsymbol{x},t) \sim \rho_{\mathcal{M}_t^{\tilde{\boldsymbol{R}}}}} \bigg( \mathcal{L}_{rec}(f_\theta(\boldsymbol{R}_\varphi^T \boldsymbol{x}, t, \boldsymbol{z}), \mathrm{SDF}_{\mathcal{M}_t}(\tilde{\boldsymbol{R}}^T \boldsymbol{x})) + \mathcal{L}_{code}(\boldsymbol{z}) \bigg). \tag{S.3}$$

Writing out the expectation in integral form, we obtain:

$$\int_0^1 \int_\Omega \mathcal{L}_{rec}(f_\theta(\boldsymbol{R}_\varphi^T \boldsymbol{x}, t, \boldsymbol{z}), \mathrm{SDF}_{\mathcal{M}_t}(\tilde{\boldsymbol{R}}^T \boldsymbol{x})) \rho_{\mathcal{M}_t^{\tilde{\boldsymbol{R}}}} \boldsymbol{x}, t) \mathrm{d}\boldsymbol{x} \mathrm{d}t + \mathcal{L}_{code}(\boldsymbol{z}) \tag{S.4}$$

1

Making the substitution $\boldsymbol{y} = \tilde{\boldsymbol{R}}^T\boldsymbol{x}$ with $\boldsymbol{x} = \tilde{\boldsymbol{R}}\boldsymbol{y}$ yields:

$$\int_0^1 \int_{\tilde{\boldsymbol{R}}^T\Omega} \mathcal{L}_{rec}(f_\theta(\boldsymbol{R}_\varphi^T\tilde{\boldsymbol{R}}\boldsymbol{y}, t, \boldsymbol{z}), \text{SDF}_{\mathcal{M}_t}(\boldsymbol{y}))\rho_{\mathcal{M}_t^{\tilde{R}}}(\tilde{\boldsymbol{R}}\boldsymbol{y}, t)\det(\tilde{\boldsymbol{R}})\mathrm{d}\boldsymbol{y}\mathrm{d}t + \mathcal{L}_{code}(\boldsymbol{z}) \tag{S.5}$$

Using that $\det(\tilde{\boldsymbol{R}}) = 1$ and $\tilde{\boldsymbol{R}}^T\Omega := \{\tilde{\boldsymbol{R}}^T\boldsymbol{x} \mid \boldsymbol{x} \in \Omega\} = \Omega$ as $\Omega$ is a sphere centered around the origin and $\tilde{\boldsymbol{R}}^T$ is a rotation matrix, we get:

$$\int_0^1 \int_\Omega \mathcal{L}_{rec}(f_\theta(\boldsymbol{R}_\varphi^T\tilde{\boldsymbol{R}}\boldsymbol{y}, t, \boldsymbol{z}), \text{SDF}_{\mathcal{M}_t}(\boldsymbol{y}))\rho_{\mathcal{M}_t^{\tilde{R}}}(\tilde{\boldsymbol{R}}\boldsymbol{y}, t)\mathrm{d}\boldsymbol{y}\mathrm{d}t + \mathcal{L}_{code}(\boldsymbol{z}) \tag{S.6}$$

Using our assumption on $\rho_{\mathcal{M}_t^{\tilde{R}}}$ gives us $\rho_{\mathcal{M}_t^{\tilde{R}}}(\tilde{\boldsymbol{R}}\boldsymbol{y}, t) = \rho_{\mathcal{M}_t}(\boldsymbol{y}, t)$. Using this and defining $\tilde{\boldsymbol{R}}_\varphi = \tilde{\boldsymbol{R}}^T\boldsymbol{R}_\varphi$ gives us:

$$\int_0^1 \int_\Omega \mathcal{L}_{rec}(f_\theta(\tilde{\boldsymbol{R}}_\varphi^T\boldsymbol{y}, t, \boldsymbol{z}), \text{SDF}_{\mathcal{M}_t}(\boldsymbol{y}))\rho_{\mathcal{M}_t}(\boldsymbol{y}, t)\mathrm{d}\boldsymbol{y}\mathrm{d}t + \mathcal{L}_{code}(\boldsymbol{z}) = \mathcal{L}_{\mathcal{M}_t}(\tilde{\boldsymbol{R}}_\varphi, \boldsymbol{z})) \tag{S.7}$$

Hence, we have $\mathcal{L}_{\mathcal{M}_t^{\tilde{R}}}(\boldsymbol{R}_\varphi, \boldsymbol{z}) = \mathcal{L}_{\mathcal{M}_t}(\tilde{\boldsymbol{R}}_\varphi, \boldsymbol{z}))$ with $\tilde{\boldsymbol{R}}_\varphi = \tilde{\boldsymbol{R}}^T\boldsymbol{R}_\varphi$. Combining this with the fact that a minimizer of $\mathcal{L}_{\mathcal{M}_t}(\tilde{\boldsymbol{R}}_\varphi, \boldsymbol{z})$ is, by assumption, given by $(\boldsymbol{R}, \boldsymbol{z})$, we get that an optimum for $\mathcal{L}_{\mathcal{M}_t^{\tilde{R}}}(\boldsymbol{R}_\varphi, \boldsymbol{z})$ is $(\tilde{\boldsymbol{R}}\boldsymbol{R}, \boldsymbol{z})$. □

**Theorem** (Equivariance of our model when dealing with a finite sample). *Assume we rotate a shape time series $\mathcal{M}_t$ using a rotation matrix $\tilde{\boldsymbol{R}}$ and obtain shape time series $\mathcal{M}_t^{\tilde{R}}$. Furthermore, assume the reconstruction for a general shape time series is obtained by minimizing Equation S.2 over $(\boldsymbol{R}_\varphi, \boldsymbol{z})$. Finally, assume that a solution to the optimization problem for $\mathcal{M}_t$ is $(\boldsymbol{R}, \boldsymbol{z})$ and that the data points for $\mathcal{M}_t^{\tilde{R}}$ are given by $\{(\boldsymbol{y}_i, t_i, s_i)\}_{i=1}^M = \{(\tilde{\boldsymbol{R}}\boldsymbol{x}_i, t_i, s_i)\}_{i=1}^M$. Then, a solution of the optimization problem for $\mathcal{M}_t^{\tilde{R}}$ is $(\tilde{\boldsymbol{R}}\boldsymbol{R}, \boldsymbol{z})$.*

*Proof.* In this case, the loss function for the problem of the rotated shape is given by:

$$\mathcal{L}_{\mathcal{M}_t^{\tilde{R}}}^M(\boldsymbol{R}_\varphi, \boldsymbol{z}) := \frac{1}{M}\sum_{i=1}^M \left(\mathcal{L}_{rec}(f_\theta(\boldsymbol{R}_\varphi^T\boldsymbol{y}_i, t_i, \boldsymbol{z}), s_i) + \mathcal{L}_{code}(\boldsymbol{z})\right) \tag{S.8}$$

Using our assumption $\boldsymbol{y}_i = \tilde{\boldsymbol{R}}\boldsymbol{x}_i$, we get:

$$\frac{1}{M}\sum_{i=1}^M \left(\mathcal{L}_{rec}(f_\theta(\boldsymbol{R}_\varphi^T\tilde{\boldsymbol{R}}\boldsymbol{x}_i, t_i, \boldsymbol{z}), s_i) + \mathcal{L}_{code}(\boldsymbol{z})\right) \tag{S.9}$$

Defining $\tilde{\boldsymbol{R}}_\varphi = \tilde{\boldsymbol{R}}^T\boldsymbol{R}_\varphi$ gives us:

$$\frac{1}{M}\sum_{i=1}^M \left(\mathcal{L}_{rec}(f_\theta(\tilde{\boldsymbol{R}}_\varphi^T\boldsymbol{x}_i, t_i, \boldsymbol{z}), s_i) + \mathcal{L}_{code}(\boldsymbol{z})\right) = \mathcal{L}_{\mathcal{M}_t}^M(\tilde{\boldsymbol{R}}_\varphi, \boldsymbol{z}) \tag{S.10}$$

Hence, we have $\mathcal{L}_{\mathcal{M}_t^{\tilde{R}}}^M(\boldsymbol{R}_\varphi, \boldsymbol{z}) = \mathcal{L}_{\mathcal{M}_t}^M(\tilde{\boldsymbol{R}}_\varphi, \boldsymbol{z})$ with $\tilde{\boldsymbol{R}}_\varphi = \tilde{\boldsymbol{R}}^T\boldsymbol{R}_\varphi$. Combining this with the fact that a minimizer of $\mathcal{L}_{\mathcal{M}_t}^M(\tilde{\boldsymbol{R}}_\varphi, \boldsymbol{z})$ is, by assumption, given by $(\boldsymbol{R}, \boldsymbol{z})$, we get that an optimum for $\mathcal{L}_{\mathcal{M}_t^{\tilde{R}}}^M(\boldsymbol{R}_\varphi, \boldsymbol{z})$ is $(\tilde{\boldsymbol{R}}\boldsymbol{R}, \boldsymbol{z})$.

□



\end{document}